\documentclass{article}

\PassOptionsToPackage{numbers, compress}{natbib}
\usepackage{natbib}
\usepackage[preprint]{neurips_2026}
\usepackage[utf8]{inputenc}
\usepackage[T1]{fontenc}
\usepackage{amsmath}
\usepackage{amssymb}
\usepackage{hyperref}
\usepackage{url}
\usepackage{booktabs}
\usepackage{graphicx}
\usepackage{multirow}
\usepackage{amsfonts}
\usepackage{nicefrac}
\usepackage{microtype}
\usepackage{tikz}
\usepackage[table]{xcolor}
\usepackage{algorithm}
\usepackage{algpseudocode}
\usepackage{fancyvrb}
\usepackage{enumitem}
\usepackage{cleveref}
\usepackage[most]{tcolorbox}
\usepackage{listings}
\usepackage{fvextra}
\usepackage{wrapfig}
\usepackage{xspace}

\newtcolorbox{promptbox}[1][]{
  enhanced,
  colback=gray!5,
  colframe=black!70,
  boxrule=0.5pt,
  arc=2pt,
  left=6pt, right=6pt, top=6pt, bottom=6pt,
  fonttitle=\bfseries,
  title=#1
}
\lstdefinestyle{tracestyle}{
  basicstyle=\small\ttfamily,
  breaklines=true,
  breakatwhitespace=false,
  breakindent=0pt,
  columns=fullflexible,
  keepspaces=true,
  showstringspaces=false,
  upquote=true,
  literate={’}{{'}}1 {‘}{{`}}1,
  frame=none,
  xleftmargin=0pt,
  aboveskip=2pt,
  belowskip=2pt
}

\newcommand{\method}{CodeThinker\xspace}

\title{Enhancing the Code Reasoning Capabilities of LLMs via Consistency-based Reinforcement Learning}

\author{
  Zhanyue Qin$^{1}$,
  Jia Feng$^{1}$,
  Yibo Lyu$^{1}$,
  Yun Peng$^{2}$,
  Dianbo Sui$^{1}$,
  Cuiyun Gao$^{1,\thanks{Cuiyun Gao is corresponding author. Email: \texttt{gaocuiyun@hit.edu.cn}}}$,
  Qing Liao$^{1}$
  \\
  $^{1}$Harbin Institute of Technology
  \quad
  $^{2}$The Chinese University of Hong Kong
  \\
  Email: johnneyqin@gmail.com
}

\begin{document}

\maketitle

\begin{abstract}
Code reasoning refers to the task of predicting the output of a program given its source code and specific inputs.
It can measure the reasoning capability of large language models (LLMs) and also benefit downstream tasks such as code generation and mathematical reasoning.
Existing work has verified the effectiveness of reinforcement learning on the task.
However, these methods design rewards solely based on final outputs or coarse-grained signals, and neglect the inherent consistency of the stepwise reasoning process in the task.
Therefore, these methods often result in sparse reward or reward hacking, which limits the full play of enhanced learning capabilities.
To alleviate these issues, we propose CodeThinker, a consistency-driven reinforcement learning framework for code reasoning. Specifically,
the CodeThinker has three key components:
1) a stepwise reasoning-aware model training module, which utilizes a consistency tracing paradigm as a template to synthesize training data that captures the stepwise reasoning process,
2) a dynamic beam sampling strategy, which aims at improving
the quality of sampled samples under a fixed sampling budget, and 3) a consistency reward mechanism that can effectively alleviate the reward hacking.
Experiments on three popular benchmarks show that CodeThinker achieves state-of-the-art performance across multiple LLMs. For instance, it outperforms the strongest baseline by 4.3\% in accuracy
when deployed on Qwen2.5-Coder-7B-Instruct.
We also validate the effectiveness of CodeThinker on downstream tasks. Results show that without additional training, CodeThinker obtains average accuracy gains of 5.33 and 3.11 percentage points on mathematical reasoning and code reasoning tasks covering 17 programming languages, respectively.
\end{abstract}

\section{Introduction}
\begin{figure}[t]
  \centering
  \includegraphics[width=1.00\textwidth]{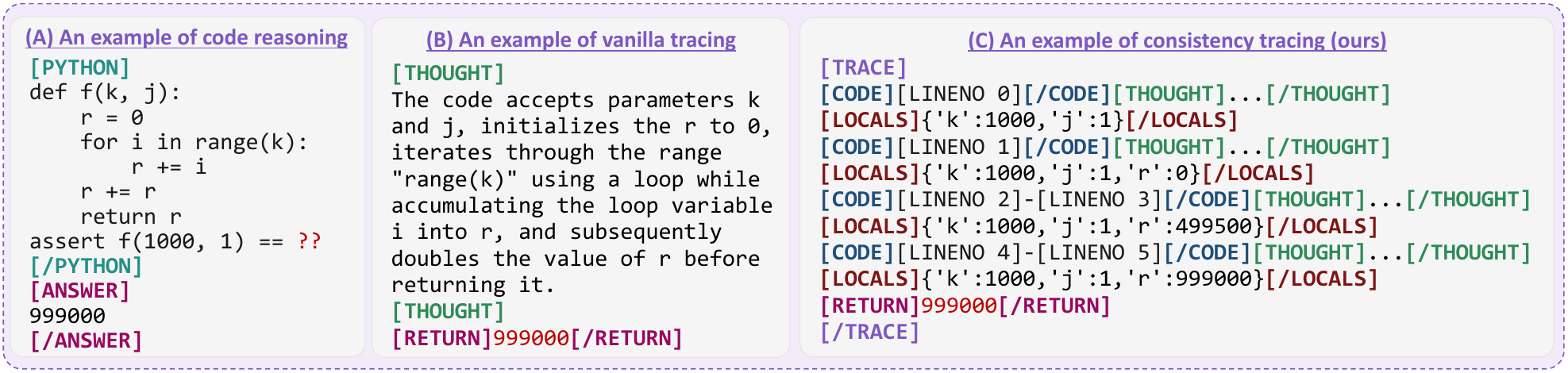}
  \caption{(A) An example of code reasoning task. (B) An example of code reasoning using vanilla tracing paradigm. (C) An example of code reasoning using our consistency tracing paradigm.}
  \label{figure:paradigm}
\end{figure}

Code reasoning (shown in \Cref{figure:paradigm}~(A)) is the task that takes a given piece of code and specific input values as conditions, analyzes the code’s execution process, and predicts the final output of this code.~\cite{gu2024cruxevalbenchmarkcodereasoning,jain2024livecodebenchholisticcontaminationfree,chen2024reasoningruntimebehaviorprogram,xu2025cruxevalxbenchmarkmultilingualcode}
It can effectively evaluate the reasoning proficiency of large language models (LLMs), while also providing substantial benefits to a wide range of downstream tasks including code generation~\cite{zhang2023repocoderrepositorylevelcodecompletion,bairi2023codeplanrepositorylevelcodingusing,pan2025catcoderrepositorylevelcodegeneration,liu2025reposcopeleveragingchainawaremultiview} and mathematical reasoning~\cite{cobbe2021trainingverifierssolvemath,gao2023palprogramaidedlanguagemodels,hendrycks2021measuringmathematicalproblemsolving,lightman2023letsverifystepstep}.

Existing work verifies the effectiveness of reinforcement learning for code reasoning~\cite{tang2025codereasonerenhancingcodereasoning,tang2026execverifywhiteboxrlverifiable}, yet mainstream methods only rely on final outputs~\cite{tang2025codereasonerenhancingcodereasoning} or coarse-grained rewards~\cite{tang2026execverifywhiteboxrlverifiable}, ignoring inherent stepwise consistency and thus suffering from sparse reward~\cite{lightman2023letsverifystepstep,yu2025dapoopensourcellmreinforcement,liu2025understandingr1zeroliketrainingcritical} and reward hacking~\cite{wu2026rewardhackingreboundsunderstanding,taylor2025schoolrewardhackshacking,denison2024sycophancysubterfugeinvestigatingrewardtampering}, which limits the improvement of LLMs’ code reasoning capability.

\begin{figure}[t]
  \centering
  \includegraphics[width=1.00\textwidth]{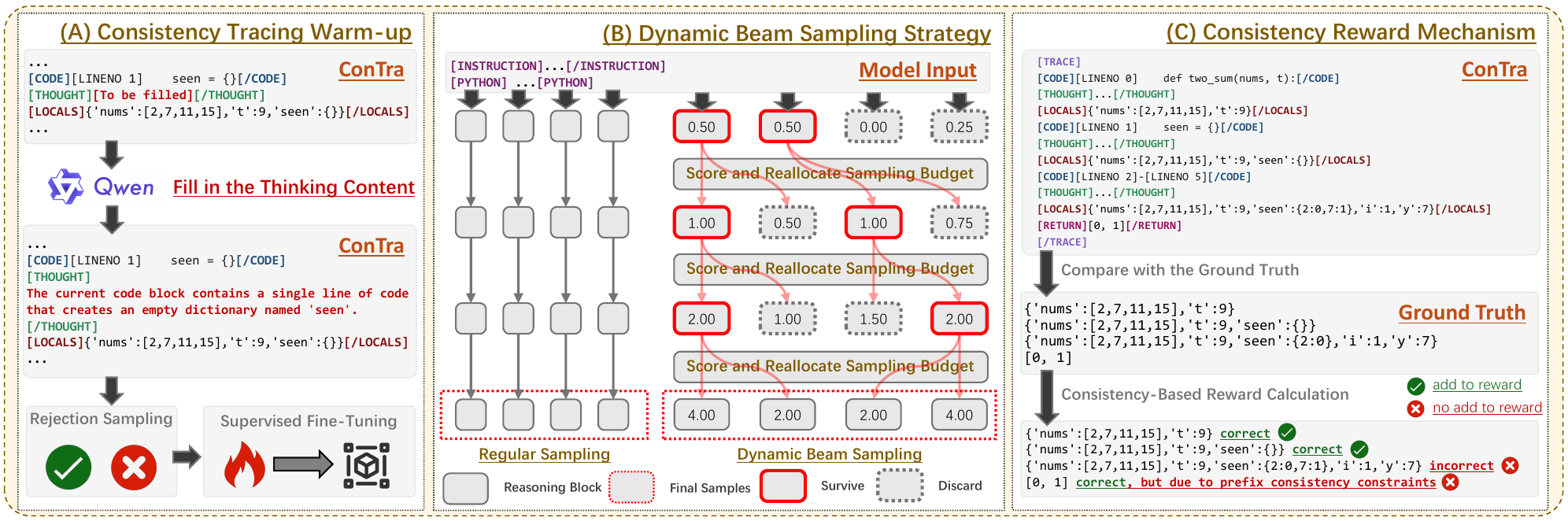}
  \caption{Workflow of the CodeThinker framework.}
  \label{figure:CodeThinker}
\end{figure}

To mitigate the issues, we propose CodeThinker framework.
It consists of the following three modules:
1) we propose a stepwise reasoning-aware model training module (\Cref{figure:CodeThinker}~(A)), which utilizes a \textbf{Con}sistency \textbf{Tra}cing (\textbf{ConTra}) paradigm (\Cref{figure:paradigm}~(C)) as a template to synthesize training data that captures the stepwise reasoning process, overcoming the limitation of vanilla tracing (\Cref{figure:paradigm}~(B)) that lacks explicit tracking for intermediate program variables.
2) Based on the consistency of the code reasoning process, we further devise a \textbf{D}ynamic \textbf{B}eam \textbf{S}ampling (\textbf{DBS}) strategy (\Cref{figure:CodeThinker}~(B)) for the sampling phase, which improves the quality of sampled examples under a non-fixed sampling budget.
3) Meanwhile, leveraging consistency in the reward calculation phase, we present a \textbf{C}onsistency \textbf{R}eward \textbf{M}echanism (\textbf{CRM}) that integrates a gated ORM and PRM (\Cref{figure:CodeThinker}~(C)).
A distinctive feature of this mechanism is that once an error occurs in the previous reasoning process, it blocks the rewards of subsequent reasoning processes, thereby solving the problem of reward hacking.


In terms of training data, we build a competition-level code reasoning training dataset called LeetCodeReasoning based on the LeetCodeDataset dataset~\cite{xia2025leetcodedatasettemporaldatasetrobust}.
Experiments demonstrate that CodeThinker achieves an average accuracy of 82.57\% across CRUXEval~\cite{gu2024cruxevalbenchmarkcodereasoning}, LiveCodeBench~\cite{jain2024livecodebenchholisticcontaminationfree}, and REval~\cite{chen2024reasoningruntimebehaviorprogram}, surpassing the best baseline CodeReasoner by 4.3\% on Qwen2.5-Coder-7B-Instruct~\cite{hui2024qwen25codertechnicalreport}.
The improvements generalize across model families: CodeThinker yields gains of 8.55, 8.54, 9.76, and 22.02 percentage points on Qwen2.5-Coder-14B/32B-Instruct, DeepSeek-Coder-6.7B-Instruct~\cite{guo2024deepseekcoderlargelanguagemodel}, and Llama-3-8B-Instruct~\cite{grattafiori2024llama3herdmodels}, respectively.
Beyond code reasoning of Python, models trained with CodeThinker exhibit a 5.33 percentage points improvement on mathematical reasoning benchmarks without any math-specific training, and exhibit a 3.11 percentage points improvement on code reasoning for 17 programming languages without any extra training.

In summary, our contributions are as follows:
\begin{itemize}[leftmargin=2em]
    \item We propose a stepwise reasoning-aware module for CodeThinker with the ConTra paradigm to explicitly capture stepwise execution beyond vanilla tracing.
    \item Guided by consistency, we propose a dynamic beam sampling strategy that allows CodeThinker to obtain better quality samples under a fixed sampling budget.
    \item Additionally, for CodeThinker, we propose aconsistency reward mechanism that fuses gated ORM and PRM, which can completely resolve the problem of reward hacking.
\end{itemize}

\newpage

\section{Code Reasoning Training Data Construction}

Existing training datasets for code reasoning suffer from insufficient logical complexity. For instance, synthetic code datasets built with Python built-in functions (e.g., \texttt{list.append()}) in prior work~\cite{tang2026execverifywhiteboxrlverifiable} generally lack elaborate logical structures, providing limited training value for code reasoning models.
To address this limitation, we construct a competition-level code reasoning training dataset named LeetCodeReasoning built upon LeetCodeDataset~\cite{xia2025leetcodedatasettemporaldatasetrobust}, which is sourced from the LeetCode algorithm competition platform.
The construction pipeline of our dataset consists of four steps.

\paragraph{Step 1: Static Filtering}
We employ Abstract Syntax Trees (AST)~\cite{ast_paper_1,ast_paper_2} to perform static analysis on the code within the LeetCodeDataset.
We retain only function-level code, filter out samples containing fewer than five lines of code, and remove redundant information such as blank lines, comments, and function signatures.
Following this step, we obtain 1,673 distinct code snippets.

\paragraph{Step 2: Unit Test Generation}
Leveraging Qwen3-32B~\cite{yang2025qwen3technicalreport}, we generate 15 sets of function inputs for each of the 1,673 functions obtained in the previous step. Since the model only needs to comprehend the function signature and input format, this generation task is straightforward and yields high-quality results. This step produces a total of 25,095 \texttt{(code, input)} pairs.

\paragraph{Step 3: Execution Filtering}
We execute each function with its corresponding generated input to obtain the expected output, forming \texttt{(code, input, output)} triples. We then apply rigorous filtering: discarding samples that failed to execute, samples that timed out (2s), and so on. Following this step, we retain 10,132 samples.

\paragraph{Step 4: Similarity Filtering}
Inspired by \citet{tang2026execverifywhiteboxrlverifiable}, we use the \texttt{all-mpnet-base-v2} sentence embedding model to vectorize the code from the 10,132 samples obtained in the previous step, as well as code from three benchmarks: CRUXEval~\cite{gu2024cruxevalbenchmarkcodereasoning}, LiveCodeBench~\cite{jain2024livecodebenchholisticcontaminationfree}, and REval~\cite{chen2024reasoningruntimebehaviorprogram}.
Subsequently, we compute the cosine similarity between our sample code and the benchmark code, and filter out all samples with a similarity score of 0.90 or higher.

After the four steps above, we obtain 9,142 samples, covering 651 distinct functions, with an average of 14.04 unit tests per function. Compared to existing datasets, our LeetCodeReasoning exhibits higher logical complexity.
Refer to \Cref{appendix:codereasoning} view the detailed comparison of our LeetCodeReasoning with existing datasets in terms of logical complexity.

\section{CodeThinker}
In this section, we first introduce consistency tracing, a novel code reasoning paradigm proposed in \Cref{subsection:contra}.
As shown in \Cref{figure:CodeThinker}, CodeThinker comprises two training stages: consistency tracing warm-up and reinforcement learning with GRPO~\cite{shao2024deepseekmathpushinglimitsmathematical}.
We use Qwen3-32B~\cite{yang2025qwen3technicalreport} to generate high-quality samples in consistency tracing format, which is used to train the LLMs with SFT (\Cref{figure:CodeThinker}~(A)).
In GRPO stage of CodeThinker, we mainly explain the proposed dynamic beam sampling strategy in \Cref{subsection:dbs}, and the proposed consistency reward mechanism in \Cref{subsection:reward}.

\subsection{Code Reasoning Paradigm: Consistency Tracing}
\label{subsection:contra}
Existing paradigms for code reasoning primarily fall into one mainstream category, which is the vanilla tracing employed by \citet{tang2025codereasonerenhancingcodereasoning} (\Cref{figure:paradigm}~(B)).
This paradigm guides LLMs to conduct explicit thinking with the entire program as the basic granularity and directly yields final results.
Nevertheless, it lacks explicit tracking of intermediate program variables during the reasoning procedure.
As a result, valuable information cannot be effectively extracted from the thinking flow to reflect execution consistency, and the reasoning process is decoupled from the real program state, which frequently leads to errors in deducing variable values.

To address the above issues, we propose a novel code reasoning paradigm named \textbf{Con}sistency \textbf{Tra}cing (\textbf{ConTra}) on the consistency.
This paradigm requires LLMs to first divide a given program into several code blocks based on the logical coherence of the outermost code.
As shown in \Cref{figure:paradigm}~(C), the outermost loop block will be treated as a separate block for only one reasoning.
After that, ConTra requires LLMs to reason through each block sequentially.
Within the ConTra paradigm, the LLMs' output must begin with \texttt{[TRACE]} and conclude with \texttt{[/TRACE]}.
The content of each reasoning block must comprise: the actual code content of the current block (enclosed within \texttt{[CODE]} and \texttt{[/CODE]}).
The explicit thinking process undertaken by the LLM regarding that block (enclosed within \texttt{[THOUGHT]} and \texttt{[/THOUGHT]}), and the current values of the program variables explicitly tracked by the LLM (enclosed within \texttt{[LOCALS]} and \texttt{[/LOCALS]}).
Once the program tracing process is complete, an assertion must be output—placed between \texttt{[RETURN]} and \texttt{[/RETURN]}, which contains the final answer derived by the LLM.

\subsection{Dynamic Beam Sampling Strategy}
\label{subsection:dbs}

\begin{algorithm}[t]
\caption{dynamic beam sampling strategy}
\label{alg:dbs}
\small
\begin{algorithmic}[1]
\Require Prompt $x$, policy $\pi_{\theta}$, beam width $n$, max stages $T_{\max}$ (default is 10), max response length $L_{\max}$
\Ensure A set of trajectories $\mathcal{Y}$ with $|\mathcal{Y}| = n$
\State $\mathcal{B} \gets \{(\epsilon,\, 0)\}^{n}$ 
       \Comment{active beam: $n$ empty partial trajectories with zero cumulative score}
\State $\mathcal{D} \gets \emptyset$ 
       \Comment{set of completed trajectories with their cumulative scores}
\State $t \gets 0$
\While{$\mathcal{B} \neq \emptyset$}
    \State $t \gets t + 1$
    \State $\widetilde{\mathcal{B}} \gets \emptyset$
    \ForAll{$(b,\, S(b)) \in \mathcal{B}$}
        \State Sample a one-block continuation $\tilde{b} \sim \pi_{\theta}(\cdot \mid x, b)$ \quad
               \Comment{decoding stops at \texttt{[/LOCALS]} or \texttt{[/TRACE]}}
        \State $S(\tilde{b}) \gets S(b) + \mathbb{R}_{\mathrm{proc}}(\tilde{b})$
        \If{\texttt{[/TRACE]} $\in \tilde{b}$ \textbf{ or } $|\tilde{b}| \ge L_{\max}$ \textbf{ or } $t = T_{\max}$}
            \State $\mathcal{D} \gets \mathcal{D} \cup \{(\tilde{b},\, S(\tilde{b}))\}$ 
                   \Comment{per-trajectory termination}
        \Else
            \State $\widetilde{\mathcal{B}} \gets \widetilde{\mathcal{B}} \cup \{(\tilde{b},\, S(\tilde{b}))\}$
        \EndIf
    \EndFor
    \If{$\widetilde{\mathcal{B}} = \emptyset$} 
        \State $\mathcal{B} \gets \emptyset$;\quad \textbf{break} 
        \Comment{global termination: no active trajectory remains}
    \EndIf
    \State $m \gets |\widetilde{\mathcal{B}}|$ 
           \Comment{current active beam size, $m \le n$}
    \State $S_{\max} \gets \max \{ S(\tilde{b}) \mid \tilde{b} \in \widetilde{\mathcal{B}} \}$
    \State $\mathcal{I} \gets \{\, \tilde{b} \in \widetilde{\mathcal{B}} \mid S(\tilde{b}) = S_{\max} \,\}$ 
           \Comment{highest-scoring active candidates}
    \State $q \gets \left\lfloor \tfrac{m}{|\mathcal{I}|} \right\rfloor$, \quad 
           $r \gets m \bmod |\mathcal{I}|$
    \State $\mathcal{B} \gets \emptyset$
    \For{$k = 1$ \textbf{to} $|\mathcal{I}|$}
        \State $c_k \gets q + \mathbb{I}[k \le r]$
        \State Add $c_k$ copies of $(\mathcal{I}_k,\, S(\mathcal{I}_k))$ to $\mathcal{B}$ 
               \Comment{winner-takes-all budget reallocation}
    \EndFor
\EndWhile
\State \Return $\mathcal{Y} \gets$ $\mathcal{D}$
\end{algorithmic}
\end{algorithm}

The regular sampling strategy in the GRPO involves providing a prompt to LLMs and having them generate multiple independent samples in a single step; however, for the relatively challenging task of code reasoning, the quality of the generated samples is generally low. To improve the upper tail of sampled trajectory quality under a fixed sampling budget, we propose a novel sampling strategy named \textbf{D}ynamic \textbf{B}eam \textbf{S}ampling (DBS), as detailed in \Cref{alg:dbs}.

Overall, unlike regular sampling in GRPO, DBS dynamically reallocates all subsequent sampling budget to the current highest-scoring partial trajectories based on intermediate PRM scores; if multiple trajectories share the highest score, the budget is divided equally among them. A schematic diagram of the dynamic beam sampling strategy is shown in \Cref{figure:CodeThinker}~(B).

\subsection{Consistency Reward Mechanism}
\label{subsection:reward}

Recent studies indicate that reinforcement learning can further enhance the code reasoning capabilities of LLMs\cite{faircodegenteam2025cwmopenweightsllmresearch,tang2026execverifywhiteboxrlverifiable}.
However, CWM~\cite{faircodegenteam2025cwmopenweightsllmresearch} and CodeReasoner~\cite{tang2025codereasonerenhancingcodereasoning} provide a pure ORM solely based on the correctness of the final answer, which leads to reward sparsity~\cite{lightman2023letsverifystepstep,yu2025dapoopensourcellmreinforcement,liu2025understandingr1zeroliketrainingcritical} and reward hacking~\cite{wu2026rewardhackingreboundsunderstanding,taylor2025schoolrewardhackshacking,denison2024sycophancysubterfugeinvestigatingrewardtampering,macdiarmid2025naturalemergentmisalignmentreward,greenblatt2024alignmentfakinglargelanguage}.
Even if ExecVerify~\cite{tang2026execverifywhiteboxrlverifiable} uses multiple white-box questions to calculate PRM, which mitigates the above issues.
However, the white-box questions in ExecVerify cannot allocate rewards to the reasoning process at a step granularity and does not take into account the consistency nature of code reasoning task.

In light of these observations, we propose a novel consistency reward mechanism specifically designed to address this consistency property of code reasoning.
Refer to \Cref{appendix:reward_design} for a detailed comparison of the consistency reward mechanism against existing work.

Let the ground-truth execution trajectory for a code-input pair be
\begin{equation}
\tau^{*} = \{(B_t^{*}, z_t^{*})\}_{t=1}^{T},
\end{equation}
where $B_t^{*}$ denotes the $t$-th code block under the ConTra code reasoning paradigm, and $z_t^{*}$ denotes the corresponding ground-truth local program state after executing that block. 
Let a model-generated trace with ConTra be
\begin{equation}
y = \{(\hat{B}_t, \hat{h}_t, \hat{z}_t)\}_{t=1}^{\hat{T}}, \hat{r}
\end{equation}
where $\hat{B}_t$, $\hat{h}_t$, and $\hat{z}_t$ denote the generated code block, explicit thinking, and tracked local state at step $t$, respectively, and $\hat{r}$ is the final predicted return value.
We define the overall reward as
\begin{equation}
\mathbb{R}(y) = \alpha \mathbb{R}_{\mathrm{proc}}(y) + \mathbb{I}\!\left[\mathbb{R}_{\mathrm{proc}}(y)=2\right] \cdot \mathbb{R}_{\mathrm{res}}(y)
\label{eq:overall_reward}
\end{equation}
where $\mathbb{R}_{\mathrm{proc}}(y)$ is the PRM score with a maximum value of 2, $\mathbb{R}_{\mathrm{res}}(y)$ is the ORM score with a maximum value of 2, and $\alpha$ balances the contribution of process-level supervision.
\paragraph{Process Reward Model (PRM)}
The PRM is designed to measure execution flow consistency. It first requires the generated trace to satisfy the ConTra format, including the correct use of \texttt{[TRACE]}, \texttt{[CODE]}, \texttt{[THOUGHT]}, \texttt{[LOCALS]}, and \texttt{[RETURN]} tags, as well as valid block boundaries and code-block alignment. Any invalid trace receives zero PRM score. For a valid trace, we further compare the generated local state with the ground-truth local state on a block-by-block basis.
Formally, let
\begin{equation}
\delta_{\mathrm{fmt}}(y) \in \{0,1\}
\end{equation}
be an indicator of whether $y$ satisfies the required ConTra format and block alignment constraints. 
For each block $t$, define the block consistency indicator
\begin{equation}
\delta_t(y) = \mathbb{I}\!\left[\hat{B}_t = B_t^{*}\right] \cdot \mathbb{I}\!\left[\hat{z}_t = z_t^{*}\right],
\label{eq:block_consistency}
\end{equation}
which is $1$ only if the generated code block and the tracked local state are both consistent with the ground-truth execution at step $t$. 
To reflect the fact that different blocks contribute unequally to the program semantics, we assign each block a nonnegative weight $w_t$, computed from the cyclomatic complexity\cite{10.5555/800253.807712} of $B_t^{*}$.
We then define the PRM as
\begin{equation}
\mathbb{R}_{\mathrm{proc}}(y)
=
2 \, \delta_{\mathrm{fmt}}(y)
\cdot
\frac{
\sum_{t=1}^{T}
w_t
\prod_{j=1}^{t}\delta_j(y)
}{
\sum_{t=1}^{T} w_t
}.
\label{eq:process_reward}
\end{equation}
This formulation captures consistency: once the generated trajectory deviates from the ground-truth execution flow at any step, subsequent blocks obtain no reward.

\paragraph{Output Reward Model (ORM)}
The ORM evaluates whether the final predicted return value matches the ground-truth program output:
\begin{equation}
\mathbb{R}_{\mathrm{res}}(y) = 2 \, \delta_{\mathrm{fmt}}(y) \cdot \mathbb{I}\!\left[\hat{r} = r^{*}\right],
\label{eq:result_reward}
\end{equation}
where $r^{*}$ denotes the ground-truth output of the program on the given input. The factor $\delta_{\mathrm{fmt}}(y)$ ensures that, just like the PRM, the ORM is granted only to traces that conform to the required ConTra format; format-violating traces are denied any credit even when their final return value happens to match the ground truth.
Importantly, as shown in \Cref{eq:overall_reward}, the ORM is activated only when the PRM reaches its maximum value, i.e., only when the entire generated trace is fully consistent with the ground-truth execution flow. 
This gating mechanism tightly couples final-answer correctness with process consistency, and prevents the model from receiving high reward merely by guessing the correct answer while producing an inconsistent intermediate trace.
\Cref{figure:CodeThinker} (C) illustrates our reward mechanism.

\newpage
\newpage

\section{Experiments}
\label{section:experiments}

In this section, we present the experimental setup in \Cref{subsection:experimental_setup}.
After that, 
we conduct extensive experiments to address the following four research questions (RQs):
\textbf{RQ1:} How effective is \method compared to baselines?
\textbf{RQ2:} What are the impacts of different modules in \method?
\textbf{RQ3:} How effective is \method on different base models?
\textbf{RQ4:} How effective is \method when generalizing to other domains?

\subsection{Experimental setup}
\label{subsection:experimental_setup}


\paragraph{Models} We adopt Qwen2.5-Coder-7B-Instruct~\cite{hui2024qwen25codertechnicalreport} (QC-7B) as the primary base model. To evaluate generalizability across model families and scales (RQ3), 
we further experiment with the 14B (QC-14B) and 32B (QC-32B) variants from the same series, as well as DeepSeek-Coder-6.7B-Instruct~\cite{guo2024deepseekcoderlargelanguagemodel} (DC-6.7B) and Llama3-8B-Instruct~\cite{grattafiori2024llama3herdmodels} (Llama3-8B).

\paragraph{\method Implementation Details}
We evenly split the LeetCodeReasoning dataset into two subsets: LeetCodeReasoning-SFT and LeetCodeReasoning-GRPO.
Unless otherwise specified, parameter $\alpha$ in \Cref{eq:overall_reward} is set to 1, and the group size of GRPO is set to 8, where 4 sampling budgets are allocated to regular sampling and the remaining 4 to DBS.
All models are trained for 3 epochs in both the SFT and GRPO stages.
Full training details are provided in \cref{appendix:training_details_codethinker}.

\paragraph{Baselines}

We compare against the following baselines:
(i) CodeIO~\cite{li2025codeiocondensingreasoningpatterns} trains LLMs using vanilla tracing data 
to predict code output from input.
(ii) SemCoder~\cite{ding2024semcodertrainingcodelanguage} 
trains LLMs with monologue-style reasoning chain-of-thought.
(iii) CodeReasoner~\cite{tang2025codereasonerenhancingcodereasoning} 
applies GRPO with ORM after SFT warm-up to improve code reasoning.
(iv) ExecVerify~\cite{tang2026execverifywhiteboxrlverifiable} introduces a verifiable white-box execution reward 
for GRPO to enhance code reasoning.
(v) CWM~\cite{faircodegenteam2025cwmopenweightsllmresearch} performs intermediate pre-training on code environment trajectories followed by multi-task RL with verifiable rewards. Since CWM is a 32B model trained from scratch, we directly compare it against QC-32B-CodeThinker.
Unless otherwise specified, all baselines share the same base models and training configurations as \method. Refer to \Cref{appendix:related_work_baselines} and \Cref{appendix:training_details_baselines} for details.

\paragraph{Benchmarks}

We evaluate on three code reasoning benchmarks:
(i) CRUXEval~\cite{li2025codeiocondensingreasoningpatterns} contains 800 samples covering both input and output prediction. We use the output prediction subtask (CRUXEval-O) in our main experiments.
(ii) LiveCodeBench\cite{ding2024semcodertrainingcodelanguage}
collects problems from LeetCode, AtCoder, and CodeForces, with 479 samples
for output prediction task: LiveCodeBench-O (LCB-O).
(iii) REval\cite{tang2025codereasonerenhancingcodereasoning}
evaluates LLMs’ ability to perform intermediate behavioral reasoning and to assess the logical consistency of their inferences.
In our main experiments, we select a subset of 651 test samples focused exclusively on function-level output prediction subtask: REval-FO. 
All experiments are repeated five times and we report the average.

\begin{wraptable}{l}{0.48\textwidth}
  \centering
  \vspace{-2.0em}
  \caption{Comparison of our \method against baselines on three mainstream benchmarks. \textbf{ZS} denotes zero-shot, \textbf{OS} denotes one-shot. The best result in each column is \textbf{bolded}, and the second-best is \underline{underlined}. All subsequent tables follow this labeling rule.}
  \vspace{0.2em} 
  \label{tab:main_results}
  \small
  \setlength{\tabcolsep}{3pt}
  \resizebox{0.48\textwidth}{!}{
  \begin{tabular}{lcccc}
  \toprule
  Model & CRUX-O & LCB-O & REval-FO & Avg. \\
  \midrule
  Org. ConTra, ZS                                                          & 60.95 & 58.08 & 67.16 & 62.06 \\
  Org. ConTra, OS                                                          & 67.03 & 63.38 & 68.85 & 66.42 \\
  Org. vanilla tracing, OS                                                  & 68.70 & 67.93 & 74.44 & 70.36 \\
  CodeIO\cite{li2025codeiocondensingreasoningpatterns}             & 68.43 & 74.24 & 77.11 & 73.26 \\
  SemCoder\cite{ding2024semcodertrainingcodelanguage}              & 70.22 & 73.03 & 76.90 & 73.38 \\
  ExecVerify\cite{tang2026execverifywhiteboxrlverifiable}          & 71.88 & 82.00 & 78.34 & 77.41 \\
  CodeReasoner\cite{tang2025codereasonerenhancingcodereasoning}    & \underline{73.45} & \underline{84.59} & \underline{79.57} & \underline{79.20} \\
  \rowcolor{gray!25}
  CodeThinker, Ours                                               & \textbf{76.88} & \textbf{86.30} & \textbf{84.52} & \textbf{82.57} \\
  \bottomrule
  \end{tabular}
  }
  \vspace{-1.0em}
\end{wraptable}

\subsection{RQ1: How Effective is CodeThinker Compared to Baselines?}



As shown in \Cref{tab:main_results}, \method achieves the best performance on all three benchmarks with an average of 82.57\%, surpassing the second-best CodeReasoner by 3.37 percentage points (pp). The gain is consistent across benchmarks (+3.43 on CRUX-O, +1.71 on LCB-O, +4.95 on REval-FO), with the largest margin on REval-FO where intermediate behavioral reasoning benefits most from our process-level reward.
Notably, RL-enhanced baselines (ExecVerify, CodeReasoner) already outperform SFT-only methods (CodeIO, SemCoder) by 4--6 pp, yet \method further improves upon CodeReasoner and ExecVerify by 3.37 and 5.16 pp, suggesting that the gains stem from our fine-grained reward decomposition and dynamic sampling rather than RL training alone.
On other models, the comparison results between CodeThinker and the two most powerful baselines—ExecVerify and CodeReasoner—are shown in \Cref{tab:appendix_more_models_comparison}.

\begin{wraptable}{l}{0.50\textwidth}
  \centering
  \vspace{-2.0em}
  \caption{Ablation study of CodeThinker on three mainstream benchmarks. \textbf{RS} denotes rejection sampling, \textbf{RL} denotes the reinforcement learning stage, \textbf{PRM} denotes process reward model, \textbf{DBS} denotes dynamic beam sampling, \textbf{RCC} denotes reward consistency constraint, \textbf{ORM} denotes outcome reward model, and \textbf{CCW} denotes cyclomatic complexity weight.}
  \vspace{0.4em} 
  \label{tab:ablation_results}
  \small
  \setlength{\tabcolsep}{3pt}
  \resizebox{0.50\textwidth}{!}{
  \begin{tabular}{lcccc}
  \toprule
  Variant Model & CRUX-O & LCB-O & REval-FO & Avg. \\
  \midrule
  \underline{w/o RS \& RL}     & 70.00 & 71.57 & 73.21 & 71.59\,(\textcolor{red!75!black}{-11.0}) \\
  \underline{w/o RL}           & 71.30 & 72.28 & 73.47 & 72.68\,(\textcolor{red!75!black}{-9.89}) \\
  \underline{w/o PRM \& DBS}    & 74.93 & 84.59 & 82.15 & 80.56\,(\textcolor{red!75!black}{-2.01}) \\
  \underline{w/o RCC \& DBS}   & 75.95 & 84.89 & 82.95 & 81.26\,(\textcolor{red!75!black}{-1.31}) \\
  \underline{w/o ORM \& DBS}    & 75.80 & 84.34 & \underline{83.07} & 81.07\,(\textcolor{red!75!black}{-1.50}) \\
  \underline{w/o CCW \& DBS}   & 74.50 & 84.80 & 82.95 & 80.75\,(\textcolor{red!75!black}{-1.82}) \\
  \underline{w/o DBS}          & \underline{76.25} & \underline{85.18} & 83.01 & \underline{81.48}\,(\textcolor{red!75!black}{-1.09}) \\
  \midrule
  \rowcolor{gray!25}
  CodeThinker                          & \textbf{76.88} & \textbf{86.30} & \textbf{84.52} & \textbf{82.57} \\
  \bottomrule
  \end{tabular}
  }
  \vspace{-2.0em}
\end{wraptable}

\subsection{RQ2: What are the Impacts of Different Modules in CodeThinker?}


We conduct ablation experiments in \Cref{tab:ablation_results}.
Removing the RL stage causes the largest drop ($-$9.89 pp), showing that reward-driven RL yields far greater gains than supervised imitation alone.
Among the RL components, removing DBS alone causes a $-$1.09 pp drop, confirming that dynamic budget reallocation improves trajectory quality. When further removing individual signals on top of w/o DBS, PRM incurs the largest additional decline ($-$0.92 pp), highlighting the value of dense process supervision; CCW ($-$0.73 pp) shows complexity-aware weighting focuses training on critical blocks; ORM ($-$0.41 pp) and RCC ($-$0.22 pp) contribute by verifying final correctness and preventing reward hacking, respectively.

\begin{wraptable}{r}{0.45\textwidth}
  \centering
  \vspace{-2.0em}
  \caption{Comparison of model performance before and after training with CodeThinker.}
  \vspace{0.1em} 
  \label{tab:generalization_results}
  \small
  \setlength{\tabcolsep}{3pt}
  \resizebox{0.45\textwidth}{!}{
  \begin{tabular}{lccccc}
  \toprule
  Method & CRUX-O & LCB-O & REval-FO & Avg. & Diff. \\
  \midrule
  \multicolumn{6}{c}{\cellcolor{blue!10}\textit{\textbf{Qwen2.5-Coder-14B-Instruct}}} \\
  Org. ConTra, OS         & 78.93 & 79.83 & 82.27 & 80.34 & - \\
  CodeThinker, w/o RL & 79.95 & 81.71 & 81.94 & 81.20 & +0.86 \\
  \rowcolor{gray!25}
  CodeThinker         & \textbf{85.75} & \textbf{92.40} & \textbf{88.51} & \textbf{88.89} & \textbf{+8.55} \\
  \midrule
  \multicolumn{6}{c}{\cellcolor{blue!10}\textit{\textbf{Qwen2.5-Coder-32B-Instruct}}} \\
  CWM                  & 79.65 & 69.35 & 76.07 & 75.02 & - \\
  Org. ConTra, OS         & 83.22 & 81.63 & 85.68 & 83.51 & - \\
  CodeThinker w/o RL & 84.93 & 87.10 & 86.73 & 86.25 & +2.74 \\
  \rowcolor{gray!25}
  CodeThinker         & \textbf{89.13} & \textbf{95.24} & \textbf{91.77} & \textbf{92.05} & \textbf{+8.54} \\
  \midrule
  \multicolumn{6}{c}{\cellcolor{blue!10}\textit{\textbf{DeepSeek-Coder-6.7B-Instruct}}} \\
  Org. ConTra, OS         & 43.25 & 26.60 & 40.98 & 36.94 & - \\
  CodeThinker w/o RL & 48.15 & 36.20 & 44.36 & 42.90 & +5.96 \\
  \rowcolor{gray!25}
  CodeThinker         & \textbf{50.23} & \textbf{41.67} & \textbf{48.20} & \textbf{46.70} & \textbf{+9.76} \\
  \midrule
  \multicolumn{6}{c}{\cellcolor{blue!10}\textit{\textbf{Llama3-8B-Instruct}}} \\
  Org. ConTra, OS         & 37.50 & 29.65 & 39.60 & 35.58 & - \\
  CodeThinker w/o RL & 50.10 & 47.77 & 56.59 & 51.49 & +15.91 \\
  \rowcolor{gray!25}
  CodeThinker         & \textbf{51.15} & \textbf{57.71} & \textbf{63.93} & \textbf{57.60} & \textbf{+22.02} \\
  \bottomrule
  \end{tabular}
  }
  \vspace{-2.0em}
\end{wraptable}

\subsection{RQ3: How Effective is CodeThinker on Different Base Models?}


\Cref{tab:generalization_results} reports CodeThinker applied to four additional base models (QC-14B, QC-32B, DS-6.7B, and Llama3-8B). CodeThinker consistently improves all models, with average gains ranging from +8.54 pp (QC-32B) to +22.02 pp (Llama3-8B). The RL stage contributes substantial improvements across all architectures (+3.80 to +7.69 pp beyond SFT alone), indicating that our reward design generalizes beyond a single model family. On QC-32B, CodeThinker (92.05\%) also substantially outperforms CWM (75.02\%), a 32B model pre-trained from scratch with RL, demonstrating the advantage of our structured trace format over general-purpose code trajectories.


\subsection{RQ4: How Effective is CodeThinker when Generalizing to other Domains?}
\label{subsection:more_benchmarks}

\begin{table}[h]
  \centering
  \caption{Generalization of CodeThinker across diverse benchmarks.}
  \label{tab:capability_generalization_results}

  \resizebox{1.0\linewidth}{!}{%
  \begin{tabular}{lccccccc cccccc}
  \toprule
   & \multicolumn{7}{c}{\textbf{Multilingual Code Output Reasoning}} & \multicolumn{6}{c}{\textbf{Non-Output Code Reasoning}} \\
  \cmidrule(lr){2-8} \cmidrule(lr){9-14}
  Model & Ruby & C++ & PHP & C\# & Bash & Rust & Avg. & CRUXEval-I & REval-S & REval-C & REval-P & REval-CO & Avg. \\
  \midrule
  QC-7B & 28.88 & 36.00 & 48.38 & 38.50 & 38.38 & 43.00 & 38.86 & 66.97 & 42.01 & 71.13 & \textbf{47.24} & \textbf{71.09} & 59.69 \\
  \rowcolor{gray!25}
  CodeThinker & \textbf{48.88} & \textbf{46.75} & \textbf{51.12} & \textbf{41.00} & \textbf{40.62} & \textbf{45.00} & \textbf{45.56(+6.70)} & \textbf{69.00} & \textbf{46.26} & \textbf{72.08} & 44.54 & 69.73 & \textbf{60.32(+0.63)} \\
  \bottomrule
  \end{tabular}}

  \vspace{0.6em}

  \resizebox{1.0\linewidth}{!}{%
  \begin{tabular}{lcccccc cccccc}
  \toprule
   & \multicolumn{6}{c}{\textbf{Mathematical Reasoning}} & \multicolumn{6}{c}{\textbf{Code Generation}} \\
  \cmidrule(lr){2-7} \cmidrule(lr){8-13}
  Model & GSM & GSM-Hard & MATH & MATH500 & AIME24 & Avg. & HE & HE+ & MBPP & MBPP+ & LCB-CG & Avg. \\
  \midrule
  QC-7B & 72.33 & 7.43 & 55.60 & 67.60 & 13.33 & 43.26 & \textbf{92.07} & \textbf{84.76} & 82.80 & \textbf{71.96} & 29.54 & 72.23 \\
  \rowcolor{gray!25}
  CodeThinker & \textbf{74.37} & \textbf{7.88} & \textbf{68.18} & \textbf{69.20} & \textbf{23.33} & \textbf{48.59(+5.33)} & \textbf{92.07} & \textbf{84.76} & \textbf{83.86} & 71.16 & \textbf{29.94} & \textbf{72.36(+0.13)} \\
  \bottomrule
  \end{tabular}}

\end{table}

To validate the generalizability of CodeThinker across benchmarks in other domains,
we select extra benchmarks from the following areas: 1) Multilingual Code Output Reasoning, 2) Non-Output Code Reasoning, 3) Mathematical Reasoning, and 4) Code Generation. 
\textbf{Notably, 
CodeThinker is not additionally trained on any of these domains.
}
These benchmarks' details can be found in \Cref{appendix:generalization_benchmarks}, and the results are presented in \Cref{tab:capability_generalization_results}.
From this table, we observe that CodeThinker achieves consistent improvements across all four domains.
Notably, in mathematical reasoning, the average improvement reaches as high as 5.33 percentage points, and the gain on multilingual code output reasoning is even more pronounced (+6.70 percentage points in Ruby and 5 other languages, +3.11 percentage points in 17 programming languages). For the complete multilingual code output reasoning evaluation results (17 programming languages), please check \Cref{tab:cruxeval_x_more_languages}.

\section{Discussion}

\begin{figure}[t]
  \centering
  \includegraphics[width=0.90\textwidth]{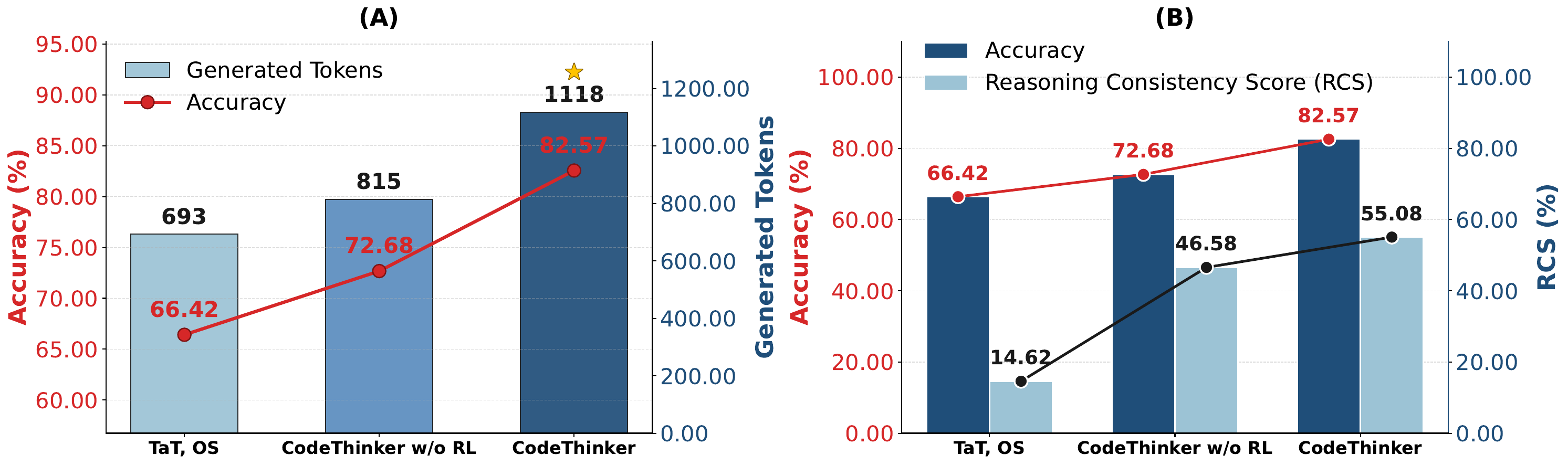}
  \caption{(A): Changes in the model's depth of thinking during the CodeThinker training. (B): Changes in the model's RCS during the CodeThinker training. \textbf{Qwen2.5-Coder-7B-Instruct}}
  \label{figure:qc_7b}
\end{figure}

\subsection{CodeThinker Deepens the LLMs' Depth of Thinking}
\label{subsection:depth_of_thinking}

We analyze the changes in the depth of model thinking of Qwen2.5-Coder-7B-Instruct before and after CodeThinker training, and the results are shown in \Cref{figure:qc_7b}(A). It can be observed that CodeThinker deepen the depth of thinking of the model, increasing from 693 tokens in the original model to 815 tokens after SFT training, and ultimately reaching 1118 tokens, while the performance also gradually improved. Additional model experimental results can be found in \Cref{appendix:more_depth_of_thinking_and_RCS}.

\subsection{Analysis of Reasoning Consistency Score}
\label{subsection:RCS}

First, we introduce the concept of the Reasoning Consistency Score (RCS). Since our ConTra reasoning paradigm is modular, we define the RCS as the proportion of correctly completed reasoning blocks relative to the total number of code blocks. (For example, if a given piece of code consists of four blocks, and the model begins to make errors starting from the third block, then the RCS is 0.5.)

We analyze the RCS of Qwen2.5-Coder-7B-Instruct on three mainstream benchmarks before and after CodeThinker training, as shown in \Cref{figure:qc_7b}(B). From the figure, we can see that after CodeThinker training, the RCS has improved from 14.62 to 55.08, which proves that CodeThinker is indeed teaching the model to perform sequential code reasoning and avoid hacking situations where the process is incorrect but the result is correct. Additional results can be found in \Cref{appendix:more_depth_of_thinking_and_RCS}.

\subsection{Dynamic Beam Sampling Improves the Quality of Sampling}

\begin{figure}[h]
  \centering
  \includegraphics[width=0.90\textwidth]{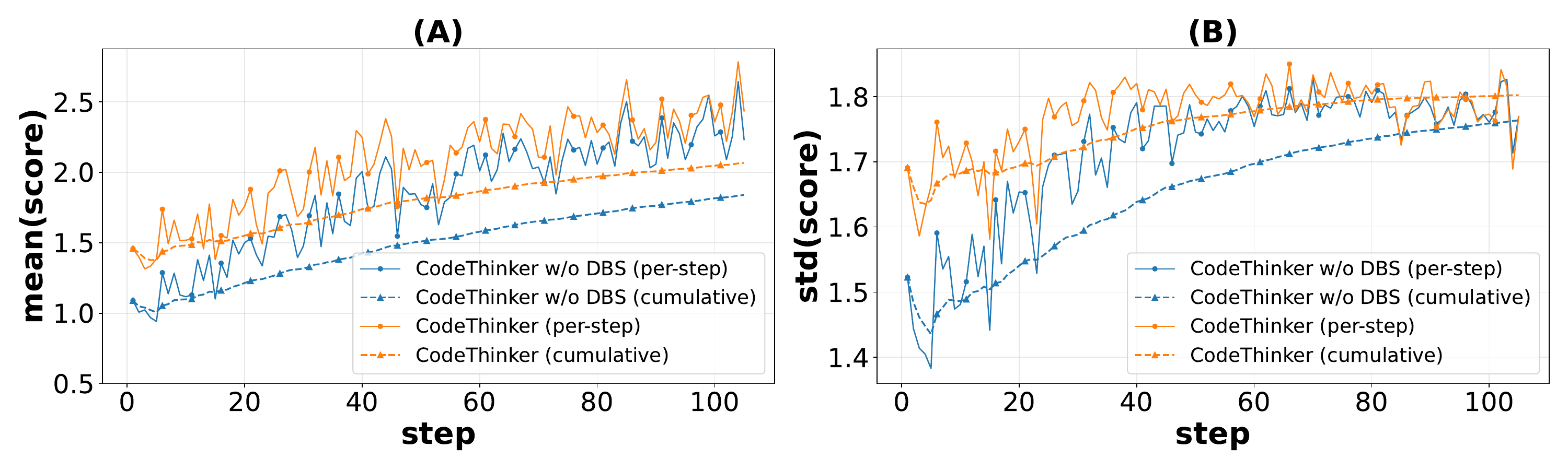}
  \caption{Changes in sample scores and standard deviations of scores during CodeThinker and CodeThinker w/o DBS training.}
  \label{figure:score_curve}
\end{figure}

To investigate whether DBS truly improves the quality of sampled samples, we analyze the sampled samples during the training of CodeThinker and CodeThinker w/o DBS, and calculate the average score and standard deviation for each step, as shown in \Cref{figure:score_curve}. From the figure, we observe that DBS indeed enhances the quality of the sampled examples, and that a higher standard deviation is beneficial to the training process.

\subsection{Impact of Regular Sampling and DBS Ratios}

\begin{wrapfigure}{r}{0.50\textwidth}
  \centering
  \vspace{-2.0em}
  \includegraphics[width=0.50\textwidth]{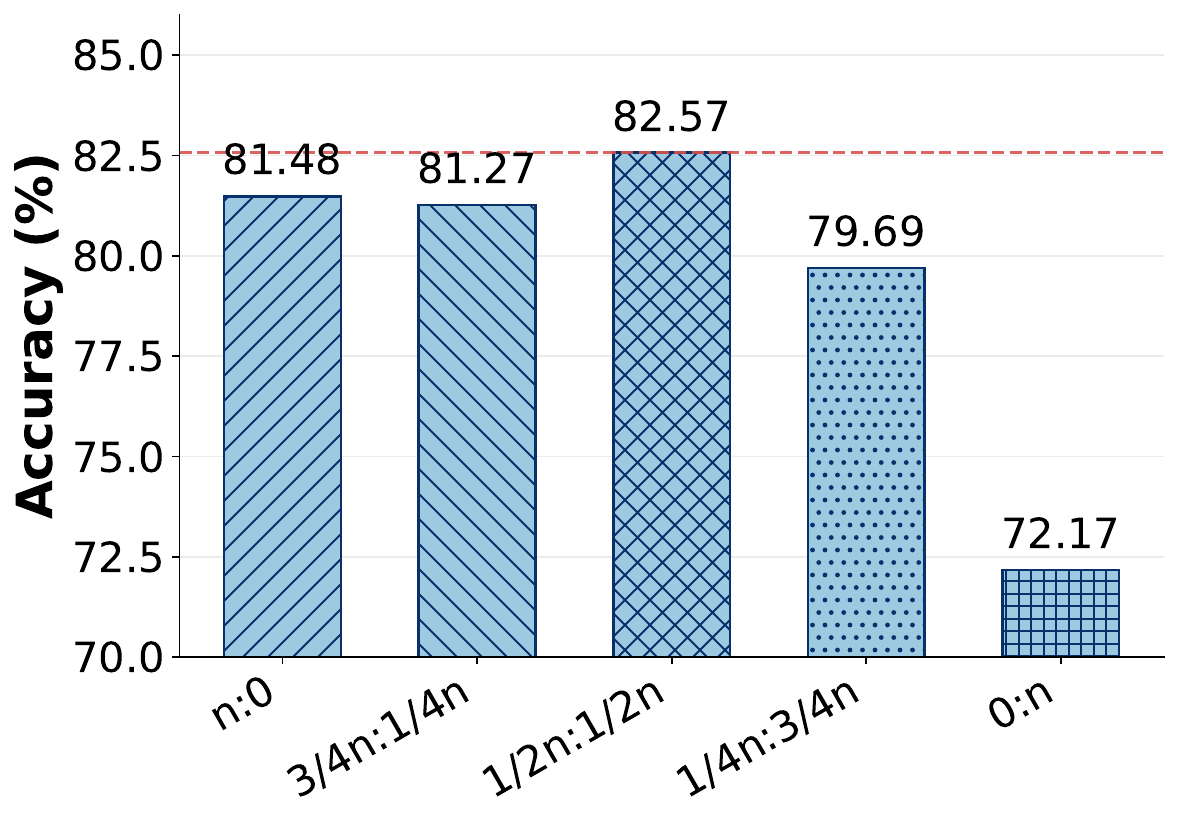}
  \caption{Results of different regular sampling and dynamic beam sampling ratios.}
  \label{figure:sampling_ratio}
  \vspace{-1.0em}
\end{wrapfigure}

To investigate the impact of the ratio between regular sampling and dynamic beam sampling within CodeThinker, we compare various sampling combinations.
The results are presented in \Cref{figure:sampling_ratio}.
The X axis of the graph represents the ratio of regular sampling to dynamic beam sampling.
From this figure, we can find clearly that CodeThinker achieves its optimal performance when the number of conventional samples and dynamic beam samples each account for half of the total sampling count.
In this configuration, conventional sampling serves to establish a quality floor for the samples, while dynamic beam sampling is responsible for elevating the upper bound of sample quality.

\subsection{More Code or More Unit Test?}

\begin{wraptable}{l}{0.45\textwidth}
  \centering
  \vspace{-2.0em}
  \caption{Comparison of training results using a dataset with half the code or half the unit tests removed versus training with the original LeetCodeReasoning dataset.}
  \vspace{0.2em} 
  \label{tab:code_vs_uts}
  \small
  \setlength{\tabcolsep}{3pt}
  \resizebox{0.45\textwidth}{!}{
  \begin{tabular}{lcccc}
  \toprule
  Stage & CRUX-O & LCB-O & REval-FO & Avg. \\
  \midrule
  \multicolumn{5}{c}{\cellcolor{blue!10}\textit{\textbf{Full Code + Full Unit Tests}}} \\
  CodeThinker w/o RL & 71.30 & 72.28 & 74.47 & 72.68 \\
  \rowcolor{gray!25}
  CodeThinker        & \textbf{76.88} & \textbf{86.30} & \textbf{84.52} & \textbf{82.57} \\
  \midrule
  \multicolumn{5}{c}{\cellcolor{blue!10}\textit{\textbf{Half Code + Full Unit Tests}}} \\
  CodeThinker w/o RL & 68.60 & 68.64 & 72.14 & 69.79 \\
  \rowcolor{gray!25}
  CodeThinker        & \underline{74.18} & \underline{80.54} & \underline{81.57} & \underline{78.76} \\
  \midrule
  \multicolumn{5}{c}{\cellcolor{blue!10}\textit{\textbf{Full Code + Half Unit Tests}}} \\
  CodeThinker w/o RL & 68.18 & 68.39 & 71.55 & 69.37 \\
  \rowcolor{gray!25}
  CodeThinker        & 73.73 & 79.87 & 79.69 & 77.76 \\
  \bottomrule
  \end{tabular}
  }
  \vspace{-2.0em}
\end{wraptable}

To analyze whether CodeThinker benefits more from a larger volume of code or from a larger volume of unit tests, we separately reduce the code and unit test components of the original LeetCodeReasoning dataset by half
We then use these processed subsets to train the model.
The results are presented in \Cref{tab:code_vs_uts}.
We can find that removing half of the code or unit tests will degrade performance.
In particular, removing half of the unit tests results in an additional 1 percentage points decline in performance compared to removing half of the code.
This demonstrates that, given a fixed budget, the model is able to derive greater result gains from more code.

\subsection{Impact of PRM and ORM ratios}

\begin{wraptable}{r}{0.50\textwidth}
  \centering
  \vspace{-2.0em}
  \caption{The impact of the ratio of PRM to ORM on CodeThinker's Performance.}
  \vspace{0.4em} 
  \label{tab:process_result}
  \small
  \setlength{\tabcolsep}{3pt}
  \resizebox{0.50\textwidth}{!}{
  \begin{tabular}{lccccc}
  \toprule
  Model & $\alpha$ & CRUX-O & LCB-O & REval-FO & Avg. \\
  \midrule
  QC-7B & $0.5$ & 76.43 & 86.47 & 83.44 & 81.21 \\
  \rowcolor{gray!25}
  QC-7B & $1$ & \textbf{76.88} & 86.30 & \textbf{84.52} & \textbf{82.57} \\
  QC-7B & $2$ & 76.10 & \textbf{86.81} & 82.37 & 81.76 \\
  \bottomrule
  \end{tabular}
  }
  \vspace{-0.0em}
\end{wraptable}

Currently, the hyperparameter $\alpha$ in the reward function (\Cref{eq:overall_reward}) of CodeThinker is set to 1.
We set $\alpha$ to 0.5 and 2, respectively, to investigate the impact of varying ratios between PRM and ORM on model performance.
The experimental results are presented in \Cref{tab:process_result}.
From this table, We can find that the best performance is achieved when $\alpha$ is set to 1.

\section{Conclusion}
Guided by consistency in code reasoning task, we propose CodeThinker.
We design a consistency-based code reasoning paradigm named ConTra, and construct a competition-level training dataset called LeetCodeReasoning.
Leveraging consistency, we put forward a sampling strategy termed DBS, which improves the quality of sampled examples without increasing the sampling budget.
In addition, we propose a reward mechanism integrating gated ORM and PRM base consistency, which can mitigate the issue of reward hacking.
After a large number of experiments, we have proved the effectiveness of CodeThinker compared to baselines.
For example, QC-7B is trained by CodeThinker and improved by 4.3\% compared to the best baselines.
We also demonstrate the effectiveness of our DBS sampling strategy (+1.09 percentage points) and discover the improvements in the ability of the trained model with CodeThinker.
In other areas such as multilingual code reasoning (+3.11 percentage points) and mathematical reasoning (+5.33 percentage points) without extra training.

\bibliographystyle{plainnat}
\bibliography{references}

\newpage

\appendix

\section{Related Work}

\subsection{Evaluating LLMs on Code Execution Reasoning Task}

\paragraph{CRUXEval~\cite{gu2024cruxevalbenchmarkcodereasoning}}
CRUXEval introduces 800 short, self-contained Python functions (3--13 lines) paired with deterministic input/output, and decomposes execution reasoning into output prediction (CRUXEval-O) and input prediction (CRUXEval-I). Functions are filtered to be solvable mentally by an undergraduate-level programmer, making CRUXEval a clean lens on the foundational execution-following capability of code LLMs.

\paragraph{LiveCodeBench~\cite{jain2024livecodebenchholisticcontaminationfree}}
LiveCodeBench addresses contamination and benchmark saturation by continuously sourcing date-tagged problems from LeetCode, AtCoder, and CodeForces, so that a model can be evaluated only on problems posted after its training cutoff. In addition to code generation, it covers self-repair, test-output prediction, and a code-execution scenario inherited from CRUXEval-O over more complex, human-written competition functions.

\paragraph{REval~\cite{chen2024reasoningruntimebehaviorprogram}}
REval explicitly probes the intermediate runtime behavior of a program, defining four tasks adapted from HumanEval and ClassEval---next-line prediction, executed-line judgment, variable-state tracing, and final-output prediction. It further introduces an Incremental Consistency score that penalizes models which give the right answer to a downstream task while failing an upstream one, putting consistency at the center of code reasoning evaluation.

\paragraph{CRUXEval-X~\cite{xu2025cruxevalxbenchmarkmultilingualcode}}
CRUXEval-X extends CRUXEval to a multilingual setting, automatically translating the original Python tasks into 19 programming languages via a test-guided ``translate--execute--repair'' pipeline that preserves content consistency across languages. The resulting benchmark contains roughly 19{,}000 test cases and reveals strong cross-language correlations as well as the often-poor cross-lingual generalization of Python-trained code LLMs.

\subsection{Enhancing LLMs' Performance on Code Execution Reasoning Task}
\label{appendix:related_work_baselines}

\paragraph{CodeIO~\cite{li2025codeiocondensingreasoningpatterns}}
CodeIO turns large-scale code corpora into a unified input/output prediction learning task, where the model is fine-tuned to predict function outputs (or feasible inputs) entirely in natural-language chain-of-thought, thus decoupling reasoning primitives---logic-flow planning, state-space search, decision-tree traversal---from code-specific syntax. A multi-turn variant, CodeIO++, further uses execution feedback to revise incorrect predictions.

\paragraph{SemCoder~\cite{ding2024semcodertrainingcodelanguage}}
SemCoder proposes monologue reasoning: a forward monologue narrates each line's effect, the executed control flow, and variable changes, while a backward monologue abstractly describes possible prior states for non-invertible operations such as sorting or aggregation. Trained on the curated PyX corpus, this scheme yields strong CRUXEval-I/O performance with only 6.7B parameters but produces highly verbose traces and lacks any machine-verifiable variable-tracking signal.

\paragraph{CWM~\cite{faircodegenteam2025cwmopenweightsllmresearch}}
The Code World Model is a 32B open-weights LLM trained from scratch with extensive mid-training on Python interpreter traces and agentic Docker trajectories, and post-trained with multi-task RL on verifiable coding, math, and software-engineering environments. Its execution-reasoning paradigm forces the model to emit explicit line-level program state after every executed statement, achieving strong long-horizon state tracking under teacher forcing but generating verbose traces that are prone to action hallucination on real programs.

\paragraph{CodeReasoner~\cite{tang2025codereasonerenhancingcodereasoning}}
CodeReasoner is a two-stage post-training framework---instruction tuning followed by GRPO reinforcement learning---that distills a teacher's CoT during SFT and then applies pure outcome rewards (final-answer correctness only) during RL to encourage concise, accurate reasoning. Its training data is synthesized around Python built-in methods, which limits algorithmic depth, and its outcome-only reward leaves intermediate reasoning steps unsupervised.

\paragraph{ExecVerify~\cite{tang2026execverifywhiteboxrlverifiable}}
ExecVerify augments the outcome reward with verifiable white-box signals: from interpreter traces it deterministically constructs control-flow questions (next-statement prediction) and data-flow questions (variable value/type prediction), and uses a weighted sum of these binary white-box rewards together with the I/O-correctness reward as the GRPO reward function. Despite its ``stepwise'' label, however, the white-box questions are scored independently and lack any prefix-consistency constraint, so the model can still be credited for inconsistent reasoning chains.

\section{Preliminaries: Group Relative Policy Optimization (GRPO)}

We briefly review the GRPO algorithm~\cite{shao2024deepseekmathpushinglimitsmathematical} that underpins the RL stage of CodeThinker, and clarify how its sampling, advantage estimation, and surrogate objective interact with our consistency reward in \Cref{eq:overall_reward}.
\paragraph{From PPO to GRPO}
Standard PPO~\cite{schulman2017proximalpolicyoptimizationalgorithms} optimizes a parameterized policy $\pi_\theta$ via a token-level clipped surrogate objective and relies on a learned value network to estimate per-token advantages. In settings where rewards are only available at the trajectory level---as is typical for verifiable reasoning, math, and code execution tasks---training the value network is both expensive and unstable. GRPO removes the value network entirely and instead estimates advantages through a group-relative normalization of rewards over multiple rollouts that share the same prompt. This yields a critic-free, memory-efficient algorithm that has empirically proven well-suited to reasoning RL post-training.
\paragraph{Group sampling}
For each prompt $x$ drawn from the training distribution $\mathcal{D}$, GRPO uses the behavior policy $\pi_{\theta_{\mathrm{old}}}$ to sample a group of $G$ trajectories
\begin{equation}
\{y^{(i)}\}_{i=1}^{G} \stackrel{\mathrm{i.i.d.}}{\sim} \pi_{\theta_{\mathrm{old}}}(\cdot \mid x),
\end{equation}
each scored by an outcome-style reward function $\mathbb{R}(y^{(i)})$. In CodeThinker, $\mathbb{R}(\cdot)$ is the consistency reward of \Cref{eq:overall_reward}; in DBS (\Cref{alg:dbs}), the same $G$ samples are drawn under our budget-preserving allocation between independent sampling and active beam expansion, which leaves the GRPO objective formally unchanged.
\paragraph{Group-relative advantage}
Given the group of trajectory-level rewards $\{r^{(i)} = \mathbb{R}(y^{(i)})\}_{i=1}^{G}$, GRPO estimates each trajectory's advantage by standardization within the group:
\begin{equation}
\hat A^{(i)} \;=\; \frac{r^{(i)} - \mathrm{mean}\!\left(\{r^{(j)}\}_{j=1}^{G}\right)}{\mathrm{std}\!\left(\{r^{(j)}\}_{j=1}^{G}\right) + \epsilon},
\label{eq:grpo_adv}
\end{equation}
where $\epsilon$ is a small constant for numerical stability. Each token of trajectory $y^{(i)}$ inherits the same group-relative advantage $\hat A^{(i)}_t \equiv \hat A^{(i)}$. Intuitively, $\hat A^{(i)}>0$ when trajectory $i$ is better than the average rollout for the same prompt, and $\hat A^{(i)}<0$ otherwise; this group-internal contrast removes the need for a separate value baseline.
\paragraph{Surrogate objective}
Let $\rho^{(i)}_t(\theta) = \pi_\theta(y^{(i)}_t \mid x, y^{(i)}_{<t}) / \pi_{\theta_{\mathrm{old}}}(y^{(i)}_t \mid x, y^{(i)}_{<t})$ denote the token-level importance ratio. GRPO maximizes the following PPO-style clipped objective, normalized within each trajectory and averaged across the group:
\begin{equation}
\begin{aligned}
\mathcal{J}_{\mathrm{GRPO}}(\theta)
&= \mathbb{E}_{x \sim \mathcal{D},\,\{y^{(i)}\} \sim \pi_{\theta_{\mathrm{old}}}}
\!\bigg[
\frac{1}{G}\sum_{i=1}^{G}\frac{1}{|y^{(i)}|}\sum_{t=1}^{|y^{(i)}|}
\min\!\Big(
\rho^{(i)}_t(\theta)\,\hat A^{(i)}, \\
&\qquad\qquad\quad
\mathrm{clip}\!\big(\rho^{(i)}_t(\theta),\,1-\varepsilon,\,1+\varepsilon\big)\,\hat A^{(i)}
\Big)
\bigg]
\;-\;\beta\,\mathbb{D}_{\mathrm{KL}}\!\left[\pi_\theta \,\|\, \pi_{\mathrm{ref}}\right].
\end{aligned}
\label{eq:grpo_obj}
\end{equation}
where $\varepsilon$ is the clipping range and $\beta$ controls a KL regularizer against a reference policy $\pi_{\mathrm{ref}}$ (typically the SFT checkpoint). The first term encourages each token of better-than-average trajectories to be up-weighted (and conversely for below-average trajectories), while the clipping operator caps the per-step policy update, preventing destructively large ratio shifts. Compared with PPO, GRPO drops the value-function term and the GAE bootstrapping, which we empirically observe to be both unnecessary and harmful when rewards are sparse and trajectory-level.
\paragraph{KL regularization choices}
GRPO admits two common ways of incorporating the KL term: (i) as an auxiliary loss with coefficient $\beta$ as in \Cref{eq:grpo_obj}; or (ii) as a penalty subtracted from the per-token reward before computing $\hat A^{(i)}$. As listed in \Cref{tab:rl-hparams}, we deliberately disable both variants so that the policy is shaped solely by our consistency reward; we also set the entropy coefficient to zero. This minimal configuration follows recent observations~\cite{liu2025understandingr1zeroliketrainingcritical,yu2025dapoopensourcellmreinforcement} that, with a sufficiently informative process reward, KL and entropy regularizers often hurt rather than help rule-based RL on code or mathematical reasoning.
\paragraph{Why GRPO suits code reasoning}
Three properties of GRPO are particularly aligned with our setting. First, the group-relative advantage in \Cref{eq:grpo_adv} converts the consistency reward into a contrastive signal that identifies, among $G$ rollouts on the same program, which ConTra trajectory is more execution-faithful---this is exactly what our prefix-consistency design is meant to expose. Second, the absence of a learned critic makes GRPO compatible with DBS, which alters the within-prompt sampling distribution in a way that would otherwise complicate value-function training. Third, the trajectory-level reward $\mathbb{R}(y)$ in \Cref{eq:overall_reward} is naturally consumed by GRPO without any reward-shaping detour, while the hard gate $\mathbb{I}[\mathbb{R}_{\mathrm{proc}}=2]$ is preserved end-to-end.

\section{Detailed Comparison of LeetCodeReasoning with Existing Datasets}
\label{appendix:codereasoning}




\Cref{fig:sample_execverify} show a example of dataset proposed by \citet{tang2026execverifywhiteboxrlverifiable}.
From this figure, we can find that the code within this dataset consists entirely of synthesized code, and its logical complexity is low, making it poor for training.
We directly evaluate Qwen2.5-Coder-7B-Instruct on the LeetCodeReasoning dataset (500 subset) and the dataset proposed by \citet{tang2026execverifywhiteboxrlverifiable} (500 subset), aiming to measure logical complexity from the perspective of difficulty. The results are presented in \cref{tab:more_datasets}.

\begin{table}[h]
  \centering
  \caption{Results using the dataset proposed by \citet{tang2026execverifywhiteboxrlverifiable} and LeetCodeReasoning.}
  \label{tab:more_datasets}
  \small
  \setlength{\tabcolsep}{6pt}
  \begin{tabular}{lccccc}
  \toprule
  Model & LeetCodeReasoning & Dataset~(\citet{tang2026execverifywhiteboxrlverifiable}) \\
  \midrule
  Qwen2.5-Coder-7B-Instruct & 48.16 & 53.40 \\
  \bottomrule
  \end{tabular}
\end{table}


\begin{figure}[htbp]
  \centering
  \includegraphics[width=1\textwidth]{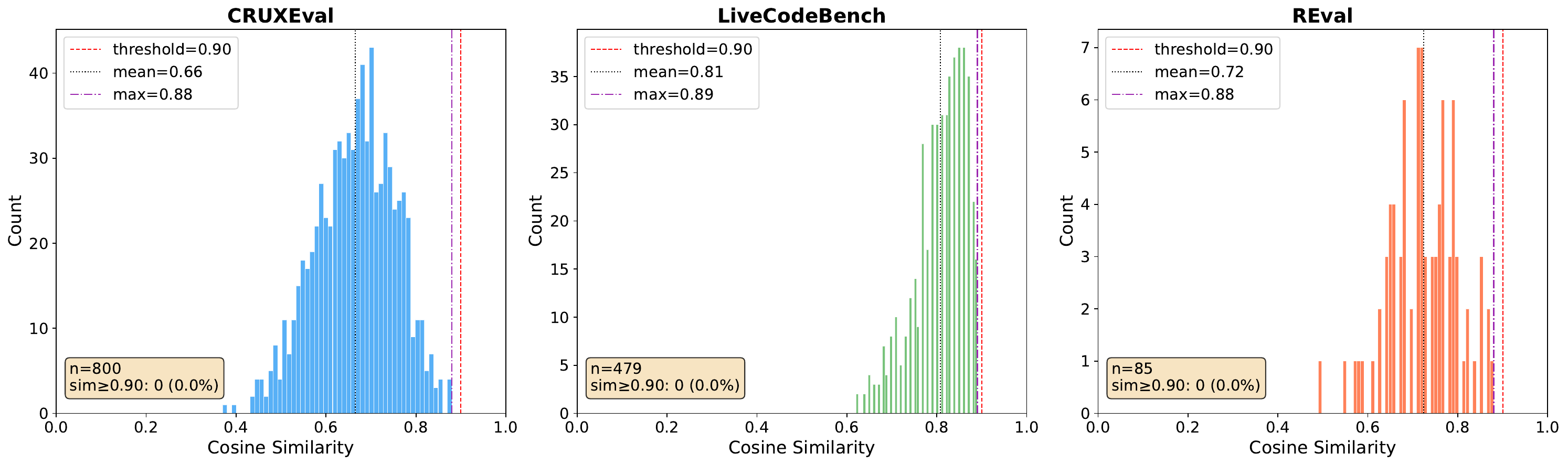}
  \caption{Cosine similarity of code in the LeetCodeReasoning dataset with code in three mainstream code reasoning benchmarks (CURXEval, LiveCodeBench, and REval).}
  \label{figure:leetcodereasoning}
\end{figure}

The prompt used to generate LeetCodeReasoning unit tests is shown in \Cref{figure:gen_uts}.
\Cref{figure:leetcodereasoning} shows the code cosine similarity distribution between our LeetCodeReasoning dataset and mainstream code reasoning benchmarks.

\section{Details of Consistency Tracing Code Reasoning Paradigm}

\subsection{Strict Formatting Requirements}

Whenever the ConTra format is required---both during rejection sampling in the SFT phase and during reward computation in the GRPO phase---we apply a strict, deterministic format validator $\delta_{\mathrm{fmt}}(\cdot)$. A trace is accepted ($\delta_{\mathrm{fmt}}=1$) only if it satisfies \emph{all} of the following constraints; otherwise it is treated as a format violation and assigned zero reward.

\paragraph{Outer wrapper.}
\begin{itemize}[leftmargin=*,nosep]
    \item The trace must begin with a single line \texttt{[TRACE]} and end with a single line \texttt{[/TRACE]}.
    \item Every tag line---\texttt{[TRACE]}, \texttt{[/TRACE]}, \texttt{[CODE]}, \texttt{[/CODE]}, \texttt{[THOUGHT]}, \texttt{[/THOUGHT]}, \texttt{[LOCALS]}, \texttt{[/LOCALS]}, \texttt{[RETURN]}, \texttt{[/RETURN]}---must appear on its own line with no leading or trailing whitespace.
\end{itemize}

\paragraph{Block structure.}
Between \texttt{[TRACE]} and \texttt{[/TRACE]}, the trace must contain one or more blocks followed by exactly one \texttt{[RETURN]} block. Each ConTra block must appear in the strict order
\[
\texttt{[CODE]} \;\rightarrow\; \texttt{[/CODE]} \;\rightarrow\; \texttt{[THOUGHT]} \;\rightarrow\; \texttt{[/THOUGHT]} \;\rightarrow\; \texttt{[LOCALS]} \;\rightarrow\; \texttt{[/LOCALS]},
\]
with no missing or reordered tags, and none of the three sections may be empty.

\paragraph{Code section.}
Every line inside \texttt{[CODE]\ldots[/CODE]} must match the pattern \texttt{[LINENO~$X$]\textvisiblespace\textvisiblespace\textvisiblespace\textvisiblespace<source line>}, where $X$ is a non-negative integer. Within a block the line numbers must be strictly consecutive ($X_{k+1}=X_k+1$); across blocks they must continue without gaps---the first block must start at \texttt{LINENO 0}, and block $i+1$ must start at $X_{\text{end}}^{(i)}+1$.

\paragraph{Locals section.}
The content of \texttt{[LOCALS]\ldots[/LOCALS]} must parse via \texttt{ast.literal\_eval} into a Python \texttt{dict}. Any unsafe expression (function calls, comprehensions over names, etc.) is rejected.

\paragraph{Return section.}
The content of \texttt{[RETURN]\ldots[/RETURN]} must be a single Python \texttt{assert} statement of the form \texttt{assert f($\cdot$) == <literal>}. The right-hand side of \texttt{==} must evaluate to a Python literal via \texttt{ast.literal\_eval} (or, as a fallback, a safe \texttt{eval}); otherwise the trace is rejected.

\paragraph{Alignment with ground truth.}
When the ground-truth execution trace is available (i.e., during rejection sampling and reward computation), the validator additionally checks: (i) the number of blocks in the model output equals the number of ground-truth blocks; (ii) each block's \texttt{LINENO} range matches the ground-truth block boundary; and (iii) the code content of each block, after stripping leading/trailing whitespace, is line-by-line identical to the ground-truth source code.

A trace satisfies $\delta_{\mathrm{fmt}}(y)=1$ iff it passes every check above; any failure short-circuits the validator and yields $\delta_{\mathrm{fmt}}(y)=0$, which in turn forces both $\mathbb{R}_{\mathrm{proc}}(y)=0$ (\Cref{eq:process_reward}) and $\mathbb{R}_{\mathrm{res}}(y)=0$ (\Cref{eq:result_reward}).

\subsection{Extraction of CoT in ConTra Format during the SFT Phase}

The prompt we used to extract the CoT in ConTra format is illustrated in \Cref{figure:extract_tat_input} and \Cref{figure:extract_tat_output}. Unlike directly instructing LLMs to generate a complete ConTra formatted reasoning chain, we merely ask the LLMs to fill in the missing \texttt{[THOUGHT]} content, thereby reducing the difficulty of the extraction task. Following the extraction, we perform rejection sampling on the generated samples based on whether they adhere to the ConTra format specifications, whether the corresponding variable values within the \texttt{[LOCALS]} section of each step are correct, and whether the final result in the \texttt{[RETURN]} section is accurate.

\subsection{An Example of ConTra Code Reasoning}

Our CodeThinker uses 0-shot ConTra code inference format. We present an code reasoning example of CodeThinker, as shown in \Cref{figure:codethinker_example}.

\section{Comparison of CodeThinker Against Existing Work in Reward Design}
\label{appendix:reward_design}

In \Cref{fig:case_codereasoner}, \Cref{fig:case_execverify}, and \Cref{fig:case_codethinker}, we respectively present a sample drawn from CodeReasoner, ExecVerify, and CodeThinker.

\paragraph{Reward analysis of the CodeReasoner rollout.}
For the rollout shown in \Cref{fig:case_codereasoner}, the CodeReasoner reward checks only two things: whether the output respects the required format (i.e., contains a \texttt{[THOUGHT]} block followed by an \texttt{[ANSWER]} block) and whether the predicted assertion equals the ground truth. Both conditions are satisfied here, so the rollout receives the maximum reward of 2. The entire intermediate reasoning chain inside \texttt{[THOUGHT]}---including any factual or logical mistakes the model may have made about variable values---is left completely unsupervised; only the final assertion contributes to the gradient. As a consequence, a rollout whose chain of thought silently drifts from the actual program state but coincidentally guesses the correct return value would be indistinguishable, in CodeReasoner's eyes, from a rollout whose intermediate reasoning is faithful step by step.

\paragraph{Reward analysis of the ExecVerify rollout.}
For the rollout shown in \Cref{fig:case_execverify}, the ExecVerify reward sums two parts. The first part rewards a correct final assertion, identical in spirit to CodeReasoner. The second part is an aggregate over a fixed-size set of independent white-box questions about intermediate execution behaviour (Q2--Q6 in this example), in which each question contributes equally to the white-box term regardless of how other questions are answered. In this rollout the assertion is correct and all five white-box answers happen to be correct as well, so the rollout earns the maximum reward of 2. Crucially, however, the white-box questions are scored independently: had the model answered, say, Q3 incorrectly while still answering Q4--Q6 correctly, it would have lost only one fifth of the white-box term and still received roughly 1.8 reward. ExecVerify therefore behaves as a multi-objective process reward---there is no consistency constraint that propagates an early mistake to subsequent steps, so an inconsistent reasoning chain can still be credited with a near-maximum reward.

\paragraph{Reward analysis of the CodeThinker rollout.}
For the rollout shown in \Cref{fig:case_codethinker}, the ground-truth execution flow---obtained by interpreting the program with the actual input---consists of four ConTra blocks $B_1, B_2, B_3, B_4$ with cyclomatic-complexity weights $w_1{=}w_2{=}w_3{=}1$ and $w_4{=}2$ (since $B_4$ contains the while loop), together with the ground-truth final value $r^{*}{=}15$. The model's output passes the strict ConTra format check ($\delta_{\mathrm{fmt}}{=}1$), and at every block the predicted code segment and the predicted local-variable state agree with the ground truth, i.e., $\delta_t{=}1$ for $t{=}1,\ldots,4$. Plugging these into the prefix-consistent process reward of \Cref{eq:process_reward} yields $\mathbb{R}_{\mathrm{proc}}(y) = 2\,\delta_{\mathrm{fmt}}\cdot\frac{\sum_t w_t\prod_{j\le t}\delta_j}{\sum_t w_t} = 2\cdot 1 \cdot \frac{5}{5} = 2$, i.e., the maximum process reward is achieved. Because the gate $\mathbb{I}[\mathbb{R}_{\mathrm{proc}}{=}2]$ in \Cref{eq:overall_reward} is now open, the result reward $\mathbb{R}_{\mathrm{res}}(y) = 2\,\delta_{\mathrm{fmt}}\cdot\mathbb{I}[\hat r{=}r^{*}] = 2$ is admitted, and the total reward becomes $\mathbb{R}(y) = \alpha\,\mathbb{R}_{\mathrm{proc}} + \mathbb{I}[\mathbb{R}_{\mathrm{proc}}{=}2]\cdot\mathbb{R}_{\mathrm{res}} = 2 + 2 = 4$ (with $\alpha{=}1$), exactly matching the score $=4$ logged in our JSONL rollout file. The qualitative difference from the previous two cases lies in the prefix-consistency factor $\prod_{j\le t}\delta_j$: had the model instead predicted $x{=}8$ in $B_3$ (a typical off-by-one bit-shift mistake), $\delta_3$ would become $0$ and the prefix product would zero out the contribution of every later block including $B_4$, dragging $\mathbb{R}_{\mathrm{proc}}$ below its maximum and closing the gate on the result reward. In such a counterfactual the rollout would receive at most a partial process reward and no result reward, even if the final return value happened to coincide with $r^{*}{=}15$ by a lucky cancellation; CodeReasoner and ExecVerify, by contrast, would still pay the full final-answer reward in this case.

\section{More Experiments}
\label{appendix:more_experiments}

\begin{table}[t]
  \centering
  \caption{Comparison of training results with different numbers of sampling rollouts during RL.}
  \label{tab:sampling_quantity}
  \resizebox{1.00\linewidth}{!}{
  \begin{tabular}{lccccc}
  \toprule
  Model & Number of samples & CRUXEval-O & LiveCodeBench-O & REval-FO & Average \\
  \midrule
  QC-7B, CodeThinker & 4  & 73.72 & 83.13 & 78.49 & 78.45 \\
  QC-7B, CodeThinker & 8  & 76.88 & 86.30 & \textbf{84.52} & 82.57 \\
  QC-7B, CodeThinker & 16 & \underline{77.20} & \underline{87.77} & 83.66 & \underline{82.88} \\
  QC-7B, CodeThinker & 32 & \textbf{77.65} & \textbf{89.35} & \underline{83.78} & \textbf{83.59} \\
  \bottomrule
  \end{tabular}
  }
\end{table}

\subsection{Impact of Sampling Quantity}

In CodeThinker, the default sampling group size is 8. We set this hyperparameter to 4, 16, and 32, respectively, to investigate its impact on model performance. The experimental results are presented in \Cref{tab:sampling_quantity}. Based on the experimental results, as the group size increases, the average performance of the model gradually improves. However, the model's performance on the REval-FO benchmark does not exhibit a gradual upward trend.

\subsection{Additional Comparisons Between CodeThinker and Baselines on More Models}
\label{appendix:more_models_comparison}

In this subsection, we present a comparison of Qwen2.5-Coder-14B-Instruct, Qwen2.5-Coder-32B-Instruct, DeepSeek-Coder-6.7B-Instruct, and Llama3-8B-Instruct—when trained using CodeThinker against the two most powerful baselines: ExecVerify and CodeReasoner. The experimental results are shown in \Cref{tab:appendix_more_models_comparison}.

From this table, we can find that: CodeThinker has advantages compared with existing baselines under different base models.

\begin{table}[t]
  \centering
  \caption{Additional comparison results between CodeThinker and baselines on more models.}
  \label{tab:appendix_more_models_comparison}
  \resizebox{1.00\linewidth}{!}{
  \begin{tabular}{llcccc}
  \toprule 	 	 	 
  Model & Method & CRUXEval-O & LiveCodeBench-O & REval-FO & Average \\
  \midrule
  \multirow{4}{*}{DeepSeek-Coder-6.7B-Instruct}
    & ExecVerify & 31.78 & 26.85 & 37.45 & 32.03 \\
    & CodeReasoner & \underline{41.58} & \underline{33.40} & \underline{46.48} & \underline{40.49} \\
    & \cellcolor{gray!25}CodeThinker & \cellcolor{gray!25}\textbf{50.23} & \cellcolor{gray!25}\textbf{41.67} & \cellcolor{gray!25}\textbf{48.20} & \cellcolor{gray!25}\textbf{46.70} \\
  \midrule 	  	 	 
  \multirow{4}{*}{Llama3-8B-Instruct}
    & ExecVerify & 40.13 & 22.38 & 38.71 & 33.74 \\
    & CodeReasoner & \underline{50.17} & \underline{49.69} & \underline{54.07} & \underline{51.31} \\
    & \cellcolor{gray!25}CodeThinker & \cellcolor{gray!25}\textbf{51.15} & \cellcolor{gray!25}\textbf{57.71} & \cellcolor{gray!25}\textbf{63.93} & \cellcolor{gray!25}\textbf{57.60} \\
  \midrule
  \multirow{4}{*}{Qwen2.5-Coder-14B-Instruct}
    & ExecVerify & 83.83 & 90.69 & 84.55 & 86.36 \\
    & CodeReasoner & \underline{85.35} & \textbf{94.07} & \underline{85.99} & \underline{88.47} \\
    & \cellcolor{gray!25}CodeThinker & \cellcolor{gray!25}\textbf{85.75} & \cellcolor{gray!25}\underline{92.40} & \cellcolor{gray!25}\textbf{88.51} & \cellcolor{gray!25}\textbf{88.89} \\
  \midrule
  Code World Model & - & 79.65 & 69.35 & 76.07 & 75.02 \\
  \midrule
  \multirow{4}{*}{Qwen2.5-Coder-32B-Instruct}
    & ExecVerify & 87.13 & 93.82 & 91.58 & 90.84 \\
    & CodeReasoner & \underline{87.53} & \underline{95.11} & \textbf{92.23} & \underline{91.62} \\
    & \cellcolor{gray!25}CodeThinker & \cellcolor{gray!25}\textbf{89.13} & \cellcolor{gray!25}\textbf{95.24} & \cellcolor{gray!25}\underline{91.77} & \cellcolor{gray!25}\textbf{92.05} \\
  \bottomrule
  \end{tabular}
  }
\end{table}

\subsection{The Depth of Thinking and RCS Variations Across Different Models}
\label{appendix:more_depth_of_thinking_and_RCS}

\begin{figure}[htbp]
  \centering
  \includegraphics[width=1\textwidth]{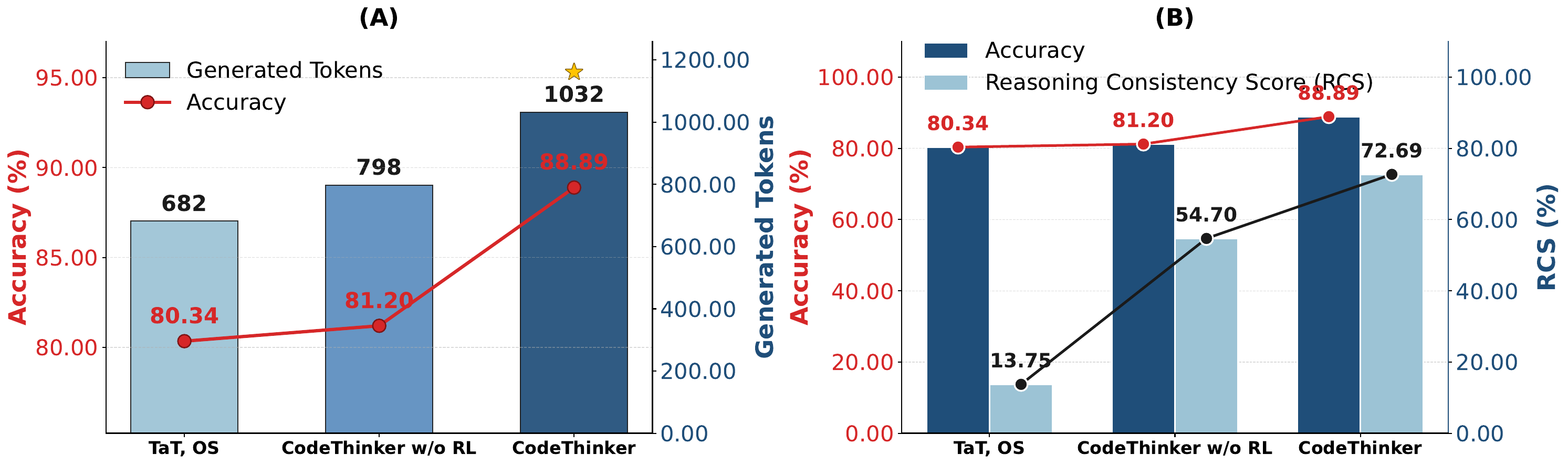}
  \caption{(A): Changes in the model's depth of thinking during the CodeThinker training. (B): Changes in the model's RCS during the CodeThinker training. \textbf{Qwen2.5-Coder-14B-Instruct}}
  \label{figure:qc_14b}
\end{figure}

\begin{figure}[htbp]
  \centering
  \includegraphics[width=1\textwidth]{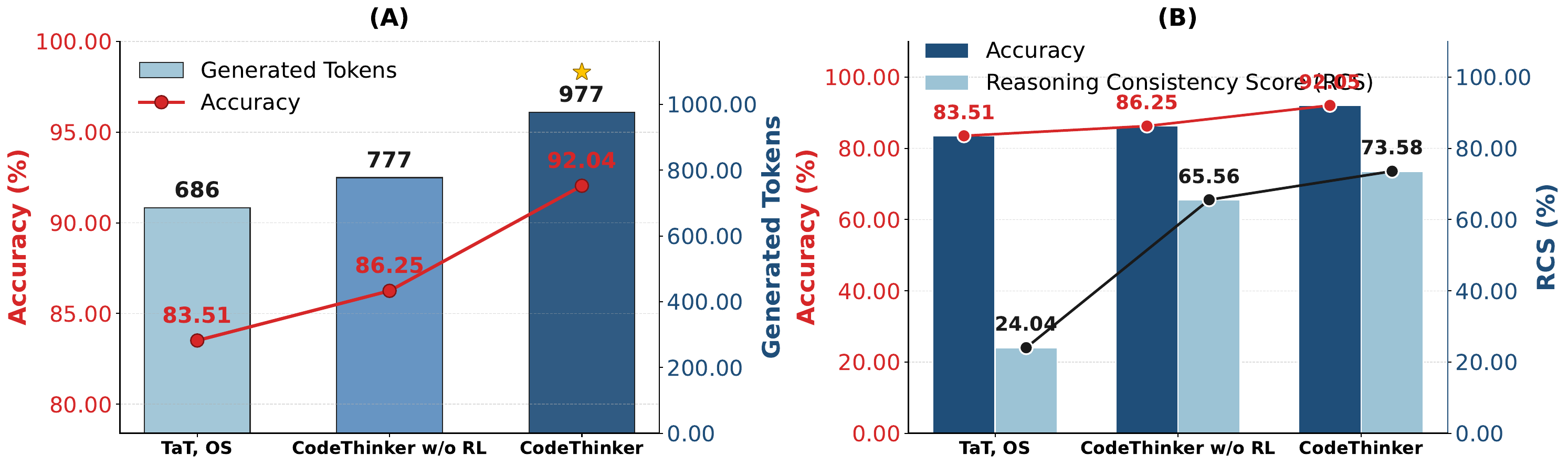}
  \caption{(A): Changes in the model's depth of thinking during the CodeThinker training. (B): Changes in the model's RCS during the CodeThinker training. \textbf{Qwen2.5-Coder-32B-Instruct}}
  \label{figure:qc_32b}
\end{figure}

\begin{figure}[htbp]
  \centering
  \includegraphics[width=1\textwidth]{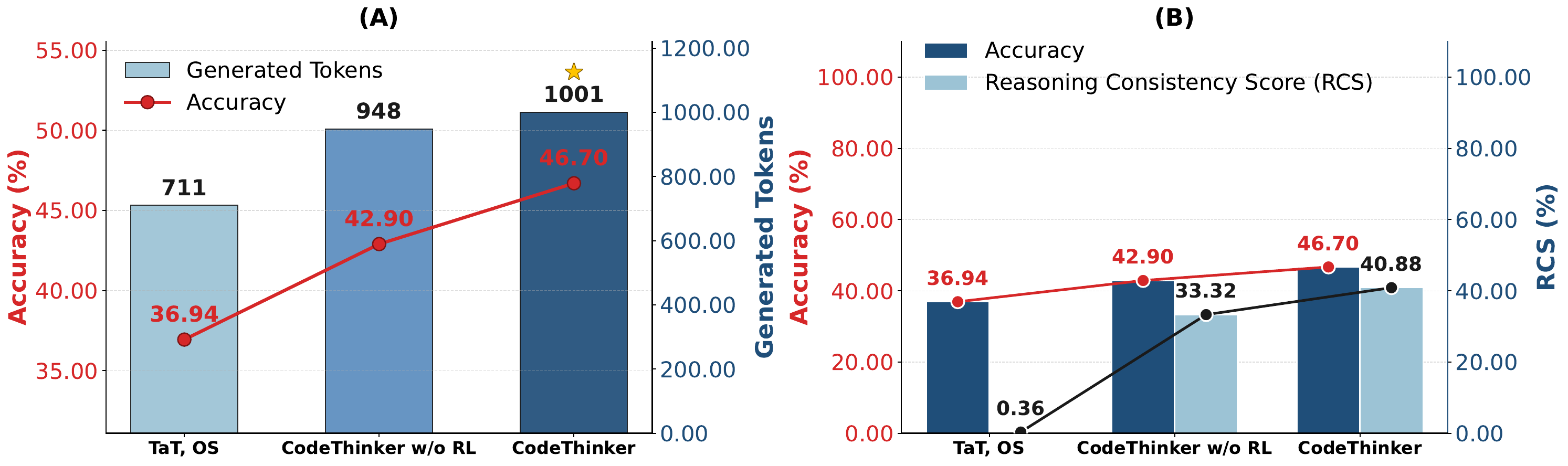}
  \caption{(A): Changes in the model's depth of thinking during the CodeThinker training. (B): Changes in the model's RCS during the CodeThinker training. \textbf{DeepSeek-Coder-6.7B-Instruct}}
  \label{figure:ds}
\end{figure}

\begin{figure}[htbp]
  \centering
  \includegraphics[width=1\textwidth]{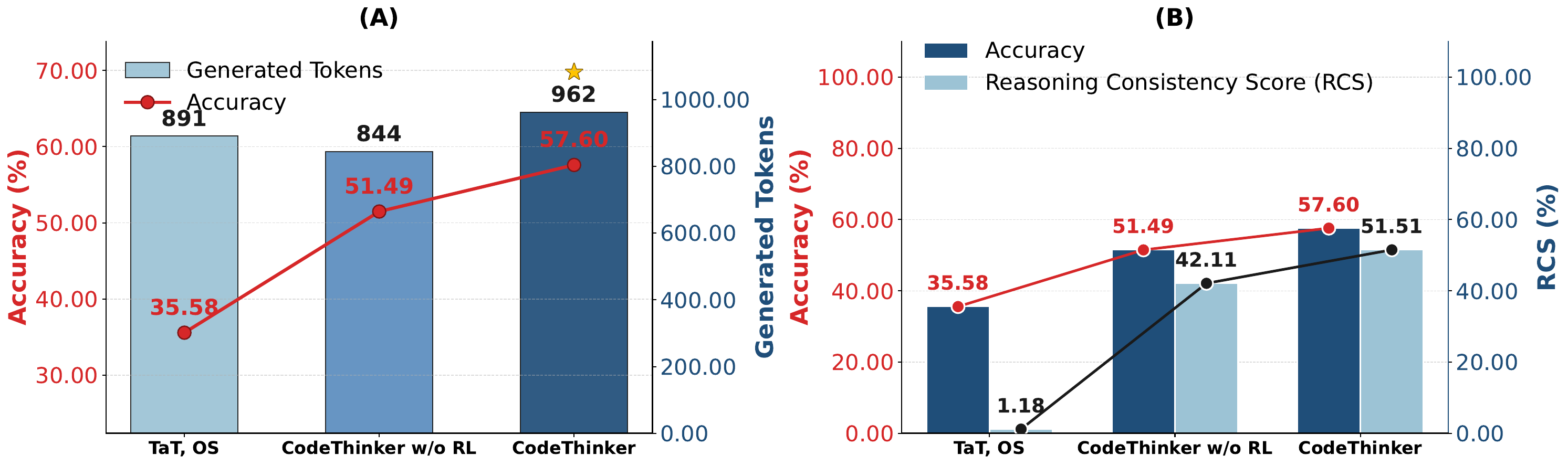}
  \caption{(A): Changes in the model's depth of thinking during the CodeThinker training. (B): Changes in the model's RCS during the CodeThinker training. \textbf{Llama3-8B-Instrcut}}
  \label{figure:llama}
\end{figure}

Employing the same experimental configuration as in \Cref{subsection:depth_of_thinking} and \Cref{subsection:RCS}, we conduct identical comparative experiments on QC-14B, QC-32B, DS-6.7B, and Llama3-8B. The experimental results are illustrated in \Cref{figure:qc_14b}, \Cref{figure:qc_32b}, \Cref{figure:ds}, and \Cref{figure:llama}. As evident from the figures, with the exception of Llama3-8B's performance regarding changes in reasoning depth, all other models demonstrated a steady increase in both reasoning depth and RCS following training with CodeThinker.

\section{Training Details of CodeThinker}
\label{appendix:training_details_codethinker}

We provide the full training hyperparameters used in the SFT and GRPO stages of the
CodeThinker. The SFT stage is implemented with the
LlamaFactory\cite{zheng2024llamafactoryunifiedefficientfinetuning} framework, and the GRPO stage is implemented with the VERL\cite{Sheng_2025} framework. All experiments are conducted on a 32×8×H20.

\subsection{SFT Training Hyperparameters}

We adopt a unified set of SFT hyperparameters across all models
(Qwen2.5-Coder-7B/14B/32B-Instruct, DeepSeek-Coder-6.7B-Instruct, and
Llama-3-8B-Instruct).
LoRA~\cite{hu2021loralowrankadaptationlarge} is used as the parameter-efficient fine-tuning method
to keep training cost consistent across model sizes.
The full configuration is summarized in \Cref{tab:sft-hparams}.

\begin{table}[h]
\centering
\caption{The full SFT hyperparameters with LlamaFactory framework.}
\label{tab:sft-hparams}
\small
\begin{tabular}{ll}
\toprule
\textbf{Hyperparameter} & \textbf{Value} \\
\midrule
fine-tuning method            & LoRA \\
LoRA rank                     & 32 \\
LoRA target modules           & all linear layers \\
maximum sequence length       & 8,192 \\
per-device batch size         & 4 \\
gradient accumulation steps   & 4 \\
pptimizer learning rate       & $5\times10^{-5}$ \\
LR scheduler                  & cosine \\
warmup ratio                  & 0.1 \\
training epochs               & 3 \\
numerical precision           & bfloat16 \\
\bottomrule
\end{tabular}
\end{table}

\subsection{GRPO Training Hyperparameters}

The GRPO stage is initialized from the SFT checkpoint and trained with GRPO.
Rollouts are produced with vLLM\cite{kwon2023efficientmemorymanagementlarge} in tensor-parallel mode under FSDP.
We deliberately disable both the KL loss term and the KL-in-reward term so
that the policy is shaped solely by the consistency-aware reward defined in
\Cref{eq:overall_reward}.
The full configuration is summarized in \Cref{tab:rl-hparams}.

\begin{table}[h]
\centering
\caption{The full GRPO hyperparameters with VERL framework.}
\label{tab:rl-hparams}
\small
\begin{tabular}{ll}
\toprule
\textbf{Hyperparameter} & \textbf{Value} \\
\midrule
advantage estimator                       & GRPO \\
maximum prompt length                     & 4,096 \\
maximum response length                   & 4,096 \\
train batch size       & 128 \\
rollouts per prompt ($n$)                 & 8 \\
PPO mini-batch size                       & 64 \\
PPO micro-batch size per GPU              & 2 \\
actor learning rate                       & $1\times10^{-6}$ \\
KL loss coefficient                       & disabled \\
KL in reward                              & disabled \\
entropy coefficient                       & 0 \\
training epochs                           & 3 \\
\midrule
rollout backend                           & vLLM \\
vLLM: tensor-parallel size              & 8 \\
vLLM: GPU memory utilization            & 0.6 \\
gradient checkpointing                    & enabled \\
numerical precision                       & bfloat16 \\
\bottomrule
\end{tabular}
\end{table}

\section{Training and Inference Details of Baselines}
\label{appendix:training_details_baselines}

In this section, we provide a detailed description of the training process of the baselines and the code inference paradigm used.

\subsection{CodeIO}

\paragraph{Training}
We process our LeetCodeReasoning dataset using the open-source data processing pipeline from CodeIO, resulting in a final set of 17,068 samples, which we then use for SFT.
The training hyperparameters employed for the SFT training are kept consistent with those used by CodeThinker during its SFT phase.

\paragraph{Inference}
CodeIO uses vanilla tracing for code reasoning. An example is shown in \Cref{figure:codeio_example}.

\subsection{SemCoder}

\paragraph{Training}
Since SemCoder's dataset processing pipeline is not open source, we extract the same number of samples from SemCoder's open source dataset as CodeIO for SFT training.
The training hyperparameters employed for the SFT training are kept consistent with those used by CodeThinker during its SFT phase.

\paragraph{Inference}
SemCoder employs its own unique code reasoning paradigm, which is similar to vanilla tracing but requires activation via the special token \texttt{[MONOLOGUE]}. An example is shown in \Cref{figure:semcoder_example}.

\subsection{CodeReasoner}

\paragraph{Training}
CodeReasoner uses SFT and GRPO for two-stage training. During the SFT stage, we use Qwen3-32B to extract the vanilla tracing of the samples in the SFT subset of LeetCodeReasoning. We perform rejection sampling on the obtained samples to filter out the bad ones. The prompt used to extract the vanilla tracing is shown in \Cref{figure:extract_cot}. During the GRPO stage, CodeReasoner employs a pure result reward mechanism: if the model's response is correct, it receives full marks; otherwise, it receives a score of zero. Consequently, we can directly utilize the ground truth answers from the GRPO subset of the LeetCodeReasoning dataset for scoring, without the need for any additional processing.
The training hyperparameters employed for the SFT and GRPO are kept consistent with those used by CodeThinker.

\paragraph{Inference}
ExecVerify uses vanilla tracing for code reasoning. An example is shown in \Cref{figure:codereasoner_example}.

\subsection{ExecVerify}

\paragraph{Training} 
ExecVerify uses SFT and GRPO for two-stage training. During the SFT phase, the fine-tuning dataset used is consistent with the dataset used by CodeReasoner during the SFT phase. During the GRPO phase, in order to calculate white-box rewards, we used their open source piplines to extract white-box questions and answers from the code in LeetCodeReasoning's GRPO subset.
The training hyperparameters employed for the SFT and GRPO are kept consistent with those used by CodeThinker.

\paragraph{Inference}
ExecVerify uses vanilla tracing for code reasoning. An example is shown in \Cref{figure:execverify_example}.

\subsection{CWM}

\paragraph{Training}
CWM is a decoder-only model architecture comprising 32 billion parameters. It is trained entirely from scratch—spanning the full pipeline from pre-training through intermediate training to post-training (SFT + RL). Consequently, it is extremely difficult to apply the specific training pipeline used for CWM to our QC-7B/14B/32B, DS-6.7B, and Llama3-8B. Therefore, we directly use CWM's publicly released weights for evaluation, comparing it in our model generalization experiments against Qwen2.5-Coder-32B-Instruct, which was trained using the CodeThinker training paradigm.

\paragraph{Inference} 
For code reasoning tasks, CWM has a unique reasoning paradigm (shown in \Cref{figure:cwm_example}. The reasoning paradigm is to let CWM execute code step by step.

\section{Details of the Extra Benchmarks in \Cref{subsection:more_benchmarks}}
\label{appendix:generalization_benchmarks}

\subsection{Multilingual Code Output Reasoning}

We utilize CRUXEval-X\cite{xu2025cruxevalxbenchmarkmultilingualcode} to evaluate the code reasoning capabilities of LLMs across multiple programming languages. This benchmark extends the Python-based CRUXEval to 19 programming languages—including C++, Java, Rust, and TypeScript—through an automated, test-driven "translate-execute-repair" pipeline. Comprising approximately 19,000 test cases (with at least 600 problems per language), it serves as a multilingual code reasoning benchmark designed to assess the cross-language generalization performance of LLMs' code reasoning abilities. We select test samples from 17 of these programming languages for our evaluation (the remaining 2 programming languages could not be assessed due to Docker corruption). The complete experimental results are presented in \Cref{tab:cruxeval_x_more_languages}, demonstrating an average improvement of 3.11 percentage points.

\begin{table}[t]
  \centering
  \caption{Complete comparative results for Qwen2.5-Coder-7B-Instruct and Qwen2.5-Coder-7B-Instruct-CodeThinker on CRUXEval-X.}
  \label{tab:cruxeval_x_more_languages}
  \resizebox{1.00\linewidth}{!}{
  \begin{tabular}{lcccccc}
  \toprule
  Model & C++ & C\# & Go & Java & Julia & JavaScript \\
  \midrule
  QC-7B & 36.00 & 38.50 & 44.88 & 27.12 & 40.88 & 47.75 \\
  \rowcolor{gray!25}
  QC-7B-CodeThinker & \textbf{46.75} & \textbf{41.00} & \textbf{45.62} & \textbf{28.75} & \textbf{41.38} & \textbf{49.00} \\
  \midrule
  Model & Lua & PHP & Python & R & Ruby & Racket \\
  \midrule
  QC-7B & 42.62 & 48.38 & 51.75 & 39.75 & 28.88 & 41.88 \\
  \rowcolor{gray!25}
  QC-7B-CodeThinker & \textbf{44.12} & \textbf{51.12} & \textbf{53.50} & \textbf{40.50} & \textbf{48.88} & \textbf{42.50} \\
  \midrule
  Model & Rust & Scala & Bash & Swift & TypeScript & Average \\
  \midrule
  QC-7B & 43.00 & 42.62 & 38.38 & 45.00 & 44.50 & 41.29 \\
  \rowcolor{gray!25}
  QC-7B-CodeThinker & \textbf{45.00} & \textbf{43.50} & \textbf{40.62} & \textbf{46.25} & \textbf{46.25} & \textbf{44.40} \\
  \bottomrule
  \end{tabular}
  }
\end{table}

\subsection{No-Output Code Reasoning}

\paragraph{CURXEval-I}
The CURXEval\cite{gu2024cruxevalbenchmarkcodereasoning} includes code reasoning tasks for code input and code output. In the main experiment, we use the output code inference subtask, CRUXEval-O. We also test the generalization ability to the CRUXEval-I subtask.

\paragraph{REval-CO, REval-P, REval-C, and REval-S}
In the main experiment, we evaluate REval-FO, which is the function-level code output inference subtask. Here, we evaluate REval-CO (the class-level code output reasoning subtask), REval-P (the model's ability to trace the next line of code to be executed in a program), REval-C (the model's ability to determine whether a specific line of code will be executed), and REval-S (the model's ability to trace program variables).

\subsection{Mathematical Reasoning}
\paragraph{GSM8K (GSM)}
GSM8K~\cite{cobbe2021trainingverifierssolvemath} consists of 8.5K linguistically diverse grade-school math word problems with natural-language step-by-step solutions and a single final numerical answer, widely used to evaluate multi-step arithmetic reasoning of LLMs.

\paragraph{GSM8K Hard (GSM-Hard)}
GSM8K-Hard~\cite{gao2023palprogramaidedlanguagemodels} is a harder variant of GSM8K obtained by replacing the original numbers with larger and less round values, so that the answer can no longer be guessed by intuition and the model must actually carry out precise computation.

\paragraph{MATH}
MATH~\cite{hendrycks2021measuringmathematicalproblemsolving} contains $12,500$ competition-style mathematics problems drawn from AMC, AIME, and other contests, spanning seven subjects (algebra, geometry, number theory, probability, etc.) and five difficulty levels, each with a complete solution.

\paragraph{MATH500}
MATH500~\cite{lightman2023letsverifystepstep} is a $500$ problem representative subset of the MATH test set introduced in PRM800K, now serving as a standard light-weight slice for evaluating mathematical reasoning and process-reward models.

\paragraph{AIME2024 (AIME24)}
AIME2024 collects $30$ official questions from the 2024 U.S. Mathematics Invitational Examinations (AIME~I and AIME~II), each with a complete answer in\([0, 999]\), and it is much more difficult than MATH benchmark.

\subsection{Code Generation}

\paragraph{HumanEval (HE)}
HumanEval~\cite{chen2021evaluatinglargelanguagemodels} consists of $164$ hand-written Python function completion questions, each question specified by a function signature and document string, and ranked by hidden unit tests.

\paragraph{MBPP}
MBPP~\cite{austin2021programsynthesislargelanguage} provides $974$ entry-level Python programming problems collected by Google, each paired with a natural-language description and three test cases covering basic list, string, and arithmetic operations.

\paragraph{HumanEval+ (HE+) and MBPP+}
HumanEval+ and MBPP+~\cite{liu2023codegeneratedchatgptreally} are reinforced versions of HumanEval and MBPP produced by EvalPlus, which expand the original test suites by roughly $80\times$ and $35\times$ respectively, thereby eliminating the spurious passes caused by under-tested cases in the original benchmarks.

\paragraph{LiveCodeBench-CG (LCB-CG)}
LiveCodeBench-CG~\cite{jain2024livecodebenchholisticcontaminationfree} is a code generation subtask of LiveCodeBench that extracts issues from LeetCode, AtCoder, and Codefences, and updates them over time to prevent training data pollution, thereby serving as a robust benchmark for evaluating the initial level coding capabilities of cutting-edge LLMs.

\section{Limitation and Future Work}

\paragraph{Language and execution environment coverage}
LeetCodeReasoning is built on top of LeetCodeDataset, so all training data is in Python.
Although our cross-language evaluation on CRUXEval-X (\Cref{tab:cruxeval_x_more_languages}) shows that CodeThinker trained purely in Python still yields consistent gains across 17 other programming languages,
we have not yet retrained on languages with runtime semantics markedly different from Python (e.g., manual memory management in C/C++, or strict static typing in Java), and whether retraining further improves performance remains open.
A natural next step is to extend CodeThinker to more language runtimes, or to port it onto a unified symbolic execution engine so that block-level states can be verified in a language-agnostic way.

\paragraph{Problem complexity and real-world code}
Constrained by LeetCodeDataset, our LeetCodeReasoning data is at the single-function level, which is still far from complex real-world, particularly repository-level code.
Mainstream code reasoning benchmarks such as CRUXEval, LiveCodeBench, and REval share the same limitation.
A promising direction is to lift code reasoning to the repository level, design evaluation protocols for repository-level code reasoning, and apply CodeThinker to such tasks.

\paragraph{Granularity in the ConTra code reasoning paradigm}
Our ConTra paradigm operates at code-block granularity, a design choice that keeps the generated execution trajectory both human-readable and machine-verifiable while controlling its length.
However, this granularity is fixed at the extraction stage: simple blocks may be over-annotated, while loops or recursive blocks with rapidly changing states may be too coarse to be fully parsed.
An adaptive ConTra variant that dynamically chooses block boundaries based on cyclomatic complexity or local data-flow density could provide finer signals where it matters and coarser signals elsewhere.

\paragraph{Beyond code execution reasoning}
We view code execution reasoning as a means to probe the depth of code understanding rather than the ultimate goal.
The reward mechanism we designed reflects an underlying consistency principle, whose applicability extends far beyond input-output prediction.
We are actively exploring this principle in other settings, including unit-test-guided code generation, fault localization, repository-level code edit prediction, and execution-trace generation for long-horizon code agents.
A comprehensive empirical study of these extensions is left to future work.

\begin{figure}[p]
\begin{promptbox}[A sample of code in dataset proposed by \citet{tang2026execverifywhiteboxrlverifiable}.]
\begin{Verbatim}[fontsize=\small,
                 breaklines=true,
                 breakanywhere=true,
                 breaksymbolleft={},
                 breaksymbolright={}]
def f(s):
    c0 = s.copy()
    s.add(3)
    c1 = s.copy()
    s.add(4)
    c2 = s.copy()
    return c0 == {1, 2} and c1 == {1, 2, 3} and (c2 == {1, 2, 3, 4}) and (c0 != c1) and (c1 != c2)
\end{Verbatim}
\end{promptbox}
\caption{A sample of code in dataset proposed by \citet{tang2026execverifywhiteboxrlverifiable}.}
\label{fig:sample_execverify}
\end{figure}

\begin{figure}[p]
\begin{promptbox}[Case 1: a CodeReasoner GRPO rollout.]
\textbf{Input:}
\begin{Verbatim}[fontsize=\footnotesize,
                 breaklines=true,
                 breakanywhere=true,
                 breaksymbolleft={},
                 breaksymbolright={}]
[SYSTEM_PROMPT]
...
[/SYSTEM_PROMPT]
[INSTRUCTION]
...
[/INSTRUCTION]
[PYTHON]
def f(s, t):
    n, m = len(s), len(t)
    i = j = 0
    rem = False
    while i < n and j < m:
        if s[i] != t[j]:
            if rem:
                break
            rem = True
        else:
            j += 1
        i += 1
    return j
assert f('abc', 'a') == ??
[/PYTHON]
\end{Verbatim}

\vspace{0.2em}
\noindent\rule{\linewidth}{0.4pt}
\vspace{0.2em}

\textbf{Output (rollout, score $=2$):}
\begin{Verbatim}[fontsize=\footnotesize,
                 breaklines=true,
                 breakanywhere=true,
                 breaksymbolleft={},
                 breaksymbolright={}]
[THOUGHT]
Let's execute the code step by step:
1. Function f is called with s = 'abc' and t = 'a'.
2. Calculate lengths: n = len(s) = 3, m = len(t) = 1.
3. Initialize indices: i = 0, j = 0.
4. Initialize rem = False.
5. Enter the while loop (i < n and j < m):
   - First iteration:
     - i = 0, j = 0
     - Check if s[i] != t[j]: 'a' != 'a' -> False
     - Since they are equal, execute `else` block: j += 1 -> j = 1
     - i += 1 -> i = 1
   - Second iteration:
     - Check if i < n (1 < 3) and j < m (1 < 1) -> False
     - Exit the loop
6. Return j, which is 1.
[/THOUGHT]
[ANSWER]
assert f('abc', 'a') == 1
[/ANSWER]
\end{Verbatim}
\end{promptbox}
\caption{Case 1: a CodeReasoner GRPO rollout.}
\label{fig:case_codereasoner}
\end{figure}

\begin{figure}[p]
\begin{promptbox}[Case 2: an ExecVerify GRPO rollout.]
\textbf{Input (abridged):}
\begin{Verbatim}[fontsize=\footnotesize,
                 breaklines=true,
                 breakanywhere=true,
                 breaksymbolleft={},
                 breaksymbolright={}]
[SYSTEM_PROMPT]
...
[/SYSTEM_PROMPT]
[INSTRUCTION]
...
[/INSTRUCTION]

Here is the code content:
1   def f(money, children):
2       if money < children:
3           return -1
4       if money > 8 * children:
5           return children - 1
6       if money == 8 * children - 4:
7           return children - 2
8       return (money - children) // 7
10  assert f(80, 16) == ????

Here are the questions (answer in the exact order listed):
Q1: Fill the assertion statement.
Q2: Which line is executed immediately after line 8 has been executed
    for the 2 time?
Q3: Which line is executed immediately after line 1 has been executed
    for the 1 time?
Q4: Which line is executed immediately after line 3 has been executed
    for the 1 time?
Q5: Which line is executed immediately after line 5 has been executed
    for the 1 time?
Q6: Which line is executed immediately after line 6 has been executed
    for the 1 time?

Format your response as <reasoning>...</reasoning><answer>...</answer>.
\end{Verbatim}

\vspace{0.2em}
\noindent\rule{\linewidth}{0.4pt}
\vspace{0.2em}

\textbf{Output (rollout, score $=2$, only the final \texttt{<answer>}
block is shown):}
\begin{Verbatim}[fontsize=\footnotesize,
                 breaklines=true,
                 breakanywhere=true,
                 breaksymbolleft={},
                 breaksymbolright={}]
<reasoning>
... step-by-step simulation of f(80, 16): the only true branch is line 8,
which returns (80 - 16) // 7 = 9. Lines 3, 5, 7 are never reached, and
line 8 is executed only once.
</reasoning>
<answer>
assert f(80, 16) == 9               # Q1, ground truth: assert f(80,16)==9
-1                                  # Q2 (line 8 executed twice? no -> -1)
if money < children:                # Q3 (after line 1 -> line 2)
-1                                  # Q4 (line 3 never executed   -> -1)
-1                                  # Q5 (line 5 never executed   -> -1)
return (money - children) // 7      # Q6 (after line 6 -> line 8)
</answer>
\end{Verbatim}
\end{promptbox}
\caption{Case 2: an ExecVerify GRPO rollout.}
\label{fig:case_execverify}
\end{figure}

\begin{figure}[p]
\begin{promptbox}[Case 3: a CodeThinker GRPO rollout.]
\textbf{Input (abridged):}
\begin{Verbatim}[fontsize=\footnotesize,
                 breaklines=true,
                 breakanywhere=true,
                 breaksymbolleft={},
                 breaksymbolright={}]
[SYSTEM_PROMPT]
...
[/SYSTEM_PROMPT]
[INSTRUCTION]
...
[/INSTRUCTION]
[PYTHON]
def f(n):
    x = 1
    while x - 1 < n:
        x <<= 1
    return x - 1
assert f(14) == ??
[/PYTHON]
\end{Verbatim}

\vspace{0.2em}
\noindent\rule{\linewidth}{0.4pt}
\vspace{0.2em}

\textbf{Output (rollout, score $=4$):}
\begin{Verbatim}[fontsize=\footnotesize,
                 breaklines=true,
                 breakanywhere=true,
                 breaksymbolleft={},
                 breaksymbolright={}]
[TRACE]
[CODE]
[LINENO 0]    def f(n):
[/CODE]
[THOUGHT] f is called with n = 14 ... [/THOUGHT]
[LOCALS] {'n': 14} [/LOCALS]

[CODE]
[LINENO 1]        x = 1
[/CODE]
[THOUGHT] Initialize x = 1. [/THOUGHT]
[LOCALS] {'n': 14, 'x': 1} [/LOCALS]

[CODE]
[LINENO 2]        while x - 1 < n:
[LINENO 3]            x <<= 1
[/CODE]
[THOUGHT]
0<14 -> x=2; 1<14 -> x=4; 3<14 -> x=8; 7<14 -> x=16;
15<14 is False, exit loop. Final x = 16.
[/THOUGHT]
[LOCALS] {'n': 14, 'x': 16} [/LOCALS]

[CODE]
[LINENO 4]        return x - 1
[/CODE]
[THOUGHT] Compute 16 - 1 = 15 and return. [/THOUGHT]
[LOCALS] {'n': 14, 'x': 16} [/LOCALS]

[RETURN]
assert f(14) == 15
[/RETURN]
[/TRACE]
\end{Verbatim}
\end{promptbox}
\caption{Case 3: a CodeThinker GRPO rollout.}
\label{fig:case_codethinker}
\end{figure}

\begin{figure}[p]
\begin{promptbox}[The prompt used in Unit Test Generation of the pipeline.]
\textbf{Input:}
\begin{Verbatim}[fontsize=\small,
                 breaklines=true,
                 breakanywhere=true,
                 breaksymbolleft={},
                 breaksymbolright={}]
[SYSTEM_PROMPT]
You are a helpful and professional code assistant.
[/SYSTEM_PROMPT]
[INSTRUCTION]
Your task is to generate corresponding function input parameters based on
a given Python function (enclosed between [PYTHON] and [/PYTHON]).
- You need to output 20 reasonable input parameters for the function.
- No need to cover special scenarios (such as boundary values).
- All the function input parameters you generate should be enclosed
  between [UTS] and [/UTS], with each line representing the input
  parameters of one function.
- Each line uses a list to store the input parameters you generate.
Do not output any other extra content. Below is an example.
[/INSTRUCTION]
[PYTHON]
def f(nums, target):
    d = {}
    for i, x in enumerate(nums):
        if (y := target - x) in d:
            return [d[y], i]
        d[x] = i
[/PYTHON]
\end{Verbatim}

\vspace{0.2em}
\noindent\rule{\linewidth}{0.4pt}
\vspace{0.2em}

\textbf{Output:}
\begin{Verbatim}[fontsize=\small,
                 breaklines=true,
                 breakanywhere=true,
                 breaksymbolleft={},
                 breaksymbolright={}]
[UTS]
[[2,7,11,15], 9]
[[3,2,4], 6]
[[3,3], 6]
[[1,2,3,4,5], 9]
[[5,7,8,9,10], 15]
[[10,20,30,40], 50]
[[1,3,5,7], 8]
[[4,8,12,16], 20]
[[9,18,27,36], 45]
[[2,4,6,8], 10]
[[11,22,33,44], 55]
[[7,14,21,28], 35]
[[1,9,18,27], 28]
[[6,12,18,24], 30]
[[8,16,24,32], 40]
[[13,26,39,52], 65]
[[5,10,15,20], 25]
[[17,34,51,68], 85]
[[3,6,9,12], 15]
[[4,7,10,13], 17]
[/UTS]
\end{Verbatim}
\end{promptbox}
\caption{The prompt used in Unit Test Generation of the pipeline.}
\label{figure:gen_uts}
\end{figure}

\begin{figure}[p]
\begin{promptbox}[The prompt used to extract CoT of ConTra during the SFT phase. \textbf{Input Part.}]
\textbf{Input:}
\begin{Verbatim}[fontsize=\scriptsize,
                 breaklines=true,
                 breakanywhere=true,
                 breaksymbolleft={},
                 breaksymbolright={}]
[SYSTEM_PROMPT]
...
[/SYSTEM_PROMPT]
[INSTRUCTION]
...
[/INSTRUCTION]
[PYTHON]
def f(nums):
    ans = [0] * len(nums)
    i, j = 0, 1
    for x in nums:
        if x > 0:
            ans[i] = x
            i += 2
        else:
            ans[j] = x
            j += 2
    return ans
assert f([1, -2, 3, -4, 5, -6]) == ??
[/PYTHON]
[RAW_TRACE]
[CODE]
[LINENO 0]    def f(nums):
[/CODE]
[THOUGHT]
# THINKING
[/THOUGHT]
[LOCALS]
{'nums': [1, -2, 3, -4, 5, -6]}
[/LOCALS]
[CODE]
[LINENO 1]        ans = [0] * len(nums)
[LINENO 2]        i, j = 0, 1
[/CODE]
[THOUGHT]
# THINKING
[/THOUGHT]
[LOCALS]
{'nums': [1, -2, 3, -4, 5, -6], 'ans': [0, 0, 0, 0, 0, 0], 'i': 0, 'j': 1}
[/LOCALS]
[CODE]
[LINENO 3]        for x in nums:
[LINENO 4]            if x > 0:
[LINENO 5]                ans[i] = x
[LINENO 6]                i += 2
[LINENO 7]            else:
[LINENO 8]                ans[j] = x
[LINENO 9]                j += 2
[/CODE]
[THOUGHT]
# THINKING
[/THOUGHT]
[LOCALS]
{'nums': [1, -2, 3, -4, 5, -6], 'ans': [1, -2, 3, -4, 5, -6], 'i': 6, 'j': 7, 'x': -6}
[/LOCALS]
[CODE]
[LINENO 10]       return ans
[/CODE]
[THOUGHT]
# THINKING
[/THOUGHT]
[LOCALS]
{'nums': [1, -2, 3, -4, 5, -6], 'ans': [1, -2, 3, -4, 5, -6], 'i': 6, 'j': 7, 'x': -6}
[/LOCALS]
[RETURN]
assert f([1, -2, 3, -4, 5, -6]) == [1, -2, 3, -4, 5, -6]
[/RETURN]
[/RAW_TRACE]
\end{Verbatim}
\end{promptbox}
\caption{The prompt used to extract CoT of ConTra during the SFT phase. \textbf{Input Part.}}
\label{figure:extract_tat_input}
\end{figure}

\begin{figure}[p]
\begin{promptbox}[The prompt used to extract CoT of ConTra during the SFT phase. \textbf{Output Part.}]
\textbf{Output:}
\begin{Verbatim}[fontsize=\scriptsize,
                 breaklines=true,
                 breakanywhere=true,
                 breaksymbolleft={},
                 breaksymbolright={}]
[TRACE]
[CODE]
[LINENO 0]    def f(nums):
[/CODE]
[THOUGHT]
Let's execute the current code block step by step:
The function f is called with the input list [1, -2, 3, -4, 5, -6]. First,
the function definition is entered, and the parameter nums is initialized
to the input list [1, -2, 3, -4, 5, -6]. No other local variables exist
yet at this stage.
[/THOUGHT]
[LOCALS]
{'nums': [1, -2, 3, -4, 5, -6]}
[/LOCALS]
[CODE]
[LINENO 1]        ans = [0] * len(nums)
[LINENO 2]        i, j = 0, 1
[/CODE]
[THOUGHT]
Let's execute the current code block step by step:
First, the current code block calculates the length of nums (which is 6),
then creates a list ans filled with 6 zeros, so ans becomes
[0, 0, 0, 0, 0, 0]. Then, initialize two variables i and j, assigning
i = 0 and j = 1 respectively. Now the local variables include nums, ans,
i, and j with their respective values.
[/THOUGHT]
[LOCALS]
{'nums': [1, -2, 3, -4, 5, -6], 'ans': [0, 0, 0, 0, 0, 0], 'i': 0, 'j': 1}
[/LOCALS]
[CODE]
[LINENO 3]        for x in nums:
[LINENO 4]            if x > 0:
[LINENO 5]                ans[i] = x
[LINENO 6]                i += 2
[LINENO 7]            else:
[LINENO 8]                ans[j] = x
[LINENO 9]                j += 2
[/CODE]
[THOUGHT]
Let's execute the current code block step by step:
The loop iterates over each element x in nums one by one:
- First  iteration (x =  1): x > 0 True,  ans[0]=1,  i=2.
- Second iteration (x = -2): x > 0 False, ans[1]=-2, j=3.
- Third  iteration (x =  3): x > 0 True,  ans[2]=3,  i=4.
- Fourth iteration (x = -4): x > 0 False, ans[3]=-4, j=5.
- Fifth  iteration (x =  5): x > 0 True,  ans[4]=5,  i=6.
- Sixth  iteration (x = -6): x > 0 False, ans[5]=-6, j=7.
After the loop finishes, local variables are nums, ans (updated), i=6,
j=7, and x=-6 (the last element in nums).
[/THOUGHT]
[LOCALS]
{'nums': [1, -2, 3, -4, 5, -6], 'ans': [1, -2, 3, -4, 5, -6], 'i': 6, 'j': 7, 'x': -6}
[/LOCALS]
[CODE]
[LINENO 10]        return ans
[/CODE]
[THOUGHT]
Let's execute the current code block step by step:
The loop has completed execution, and now the function reaches the return
statement. The value of the ans list ([1, -2, 3, -4, 5, -6]) is prepared
to be returned. All local variables remain unchanged from the end of the
loop.
[/THOUGHT]
[LOCALS]
{'nums': [1, -2, 3, -4, 5, -6], 'ans': [1, -2, 3, -4, 5, -6], 'i': 6, 'j': 7, 'x': -6}
[/LOCALS]
[RETURN]
assert f([1, -2, 3, -4, 5, -6]) == [1, -2, 3, -4, 5, -6]
[/RETURN]
[/TRACE]
\end{Verbatim}
\end{promptbox}
\caption{The prompt used to extract CoT of ConTra during the SFT phase. \textbf{Output Part.}}
\label{figure:extract_tat_output}
\end{figure}

\begin{figure}[p]
\begin{promptbox}[An example of the code reasoning of CodeThinker.]
\textbf{Input:}
\begin{Verbatim}[fontsize=\small,
                 breaklines=true,
                 breakanywhere=true,
                 breaksymbolleft={},
                 breaksymbolright={}]
[SYSTEM_PROMPT]
...
[/SYSTEM_PROMPT]
[INSTRUCTION]
...
[/INSTRUCTION]
[PYTHON]
def f(words, x):
    return [i for i, w in enumerate(words) if x in w]
assert f(['abc', 'bcd', 'aaaa', 'cbc'], 'a') == ??
[/PYTHON]
[TRACE]
\end{Verbatim}

\vspace{0.2em}
\noindent\rule{\linewidth}{0.4pt}
\vspace{0.2em}

\textbf{Output:}
\begin{Verbatim}[fontsize=\small,
                 breaklines=true,
                 breakanywhere=true,
                 breaksymbolleft={},
                 breaksymbolright={}]
[TRACE]
[CODE]
[LINENO 0]    def f(words, x):
[/CODE]
[THOUGHT]
Let's execute the current code block step by step:
The function f is called with words=['abc', 'bcd', 'aaaa', 'cbc'] and x='a'. The parameters words and x are initialized to these values. No other local variables exist yet.
[/THOUGHT]
[LOCALS]
{'words': ['abc', 'bcd', 'aaaa', 'cbc'], 'x': 'a'}
[/LOCALS]
[CODE]
[LINENO 1]        return [i for i, w in enumerate(words) if x in w]
[/CODE]
[THOUGHT]
Let's execute the current code block step by step:
The list comprehension iterates over each element in words with its index:
1. **First iteration (i=0, w='abc')**:
   - Check if 'a' in 'abc' -> True.
   - Include index 0 in the result list.
2. **Second iteration (i=1, w='bcd')**:
   - Check if 'a' in 'bcd' -> False.
   - Exclude index 1.
3. **Third iteration (i=2, w='aaaa')**:
   - Check if 'a' in 'aaaa' -> True.
   - Include index 2 in the result list.
4. **Fourth iteration (i=3, w='cbc')**:
   - Check if 'a' in 'cbc' -> False.
   - Exclude index 3.
The final result list is [0, 2]. Local variables now include i=3 and w='cbc' (from the last iteration).
[/THOUGHT]
[LOCALS]
{'words': ['abc', 'bcd', 'aaaa', 'cbc'], 'x': 'a', 'i': 3, 'w': 'cbc'}
[/LOCALS]
[RETURN]
assert f(['abc', 'bcd', 'aaaa', 'cbc'], 'a') == [0, 2]
[/RETURN]
[/TRACE]
\end{Verbatim}
\end{promptbox}
\caption{An example of the code reasoning of CodeThinker.}
\label{figure:codethinker_example}
\end{figure}

\begin{figure}[p]
\begin{promptbox}[An example of the code reasoning of CodeIO.]
\textbf{Input:}
\begin{Verbatim}[fontsize=\small,
                 breaklines=true,
                 breakanywhere=true,
                 breaksymbolleft={},
                 breaksymbolright={}]
[SYSTEM_PROMPT]
...
[/SYSTEM_PROMPT]
[INSTRUCTION]
...
[/INSTRUCTION]
[PYTHON]
def f(words, x):
    res = []
    for i, w in enumerate(words):
        if x in w:
            res.append(i)
    return res

assert f(['leet', 'code'], 'e') == ??
[/PYTHON]
[THOUGHT]
\end{Verbatim}

\vspace{0.2em}
\noindent\rule{\linewidth}{0.4pt}
\vspace{0.2em}

\textbf{Output:}
\begin{Verbatim}[fontsize=\small,
                 breaklines=true,
                 breakanywhere=true,
                 breaksymbolleft={},
                 breaksymbolright={}]
Let's execute the code step by step:

Initialize `res = []`.
Iterate through `words = ['leet', 'code']` with their indices:
- Index 0, word 'leet': 'e' is in 'leet'. Append 0 to `res`. Now `res = [0]`.
- Index 1, word 'code': 'e' is in 'code'. Append 1 to `res`. Now `res = [0, 1]`.

Return `res`, which is `[0, 1]`.

[ANSWER]
assert f(['leet', 'code'], 'e') == [0, 1]
[/ANSWER]
\end{Verbatim}
\end{promptbox}
\caption{An example of the code reasoning of CodeIO. \textbf{\texttt{[correct]}}}
\label{figure:codeio_example}
\end{figure}

\begin{figure}[p]
\begin{promptbox}[An example of the code reasoning of SemCoder.]
\textbf{Input:}
\begin{Verbatim}[fontsize=\scriptsize,
                 breaklines=true,
                 breakanywhere=true,
                 breaksymbolleft={},
                 breaksymbolright={}]
[SYSTEM_PROMPT]
...
[/SYSTEM_PROMPT]
[INSTRUCTION]
...
[/INSTRUCTION]
[PYTHON]
def f(words, x):
    l = []
    for i in range(len(words)):
        if x in words[i]:
            l.append(i)
    return l

assert f(['leet', 'code'], 'e') == ??
[/PYTHON]
[MONOLOGUE]
\end{Verbatim}

\vspace{0.2em}
\noindent\rule{\linewidth}{0.4pt}
\vspace{0.2em}

\textbf{Output:}
\begin{Verbatim}[fontsize=\scriptsize,
                 breaklines=true,
                 breakanywhere=true,
                 breaksymbolleft={},
                 breaksymbolright={}]
[MONOLOGUE]
### Understanding the Function
The function `f` takes two parameters: a list of strings (`words`) and a character (`x`). Its purpose is to find the indices of all the strings in the list that contain the specified character `x`. Here's how it works:
- It initializes an empty list `l` to store the indices of the matching strings.
- It iterates over the indices of the `words` list using a `for` loop.
- For each index `i`, it checks if the character `x` is present in the string at that index (`words[i]`).
- If `x` is found, the index `i` is appended to the list `l`.
- Finally, the function returns the list `l` containing all the indices where `x` was found.

### Key Properties and Constraints
- The input `words` is expected to be a list of strings.
- The input `x` is a single character (string of length 1).
- The function returns a list of indices, which can be empty if no strings contain the character `x`.
- The indices in the returned list correspond to the order of appearance in the original list.

### Execution Simulation
Let's simulate the execution of `f(['leet', 'code'], 'e')` step by step.
- The function is called with `words = ['leet', 'code']` and `x = 'e'`.
- An empty list `l` is initialized: `l = []`.
- The `for` loop begins, iterating over the range of the length of `words`, which is 2 (since there are two elements in the list).

**First Iteration (i = 0):**
- The condition `if x in words[i]` translates to `if 'e' in 'leet'`.
- This evaluates to `True` because 'e' is present in the string 'leet'.
- The index `0` is appended to the list `l`: `l = [0]`.

**Second Iteration (i = 1):**
- The condition `if x in words[i]` translates to `if 'e' in 'code'`.
- This evaluates to `True` because 'e' is also present in the string 'code'.
- The index `1` is appended to the list `l`: `l = [0, 1]`.

After completing the iterations, the function reaches the return statement and returns the list `l`.

### Conclusion
The function returns `[0, 1]`, indicating that both strings in the list contain the character 'e'.

Therefore, the correct assertion would be:
[/MONOLOGUE]
[ANSWER]
assert f(['leet', 'code'], 'e') == [0, 1]
[/ANSWER]
\end{Verbatim}
\end{promptbox}
\caption{An example of the code reasoning of SemCoder. \textbf{\texttt{[correct]}}}
\label{figure:semcoder_example}
\end{figure}

\begin{figure}[p]
\begin{promptbox}[An example of extracting vanilla tracing.]
\textbf{Input:}
\begin{Verbatim}[fontsize=\scriptsize,
                 breaklines=true,
                 breakanywhere=true,
                 breaksymbolleft={},
                 breaksymbolright={}]
[SYSTEM_PROMPT]
...
[/SYSTEM_PROMPT]
Your task is to complete the step-by-step code execution reasoning (marked
with "# THINKING") based on the given Python function and the incomplete assertion with specific input parameter values (enclosed in "[PYTHON]"
and "[/PYTHON]"). The expected output format is provided in "[RAW_COT]" and "[/RAW_COT]" tags. You need to:
1. Replace the "# THINKING" marker with your detailed step-by-step
   execution reasoning.
2. Keep the "[ANSWER]" block exactly unchanged.
3. Output ONLY the completed result enclosed in "[COT]" and "[/COT]"
   tags. Do NOT output any other extra information outside the specified
   tags.
Your reasoning in "[THOUGHT]" should:
- Execute the code line by line with the given input values.
- Track how variables change at each step.
- For loops, trace each iteration explicitly.
- For conditionals, explain which branch is taken and why.
- Conclude with the final return value.
[PYTHON]
def f(nums):
    ans = [0] * len(nums)
    i, j = 0, 1
    for x in nums:
        if x > 0:
            ans[i] = x
            i += 2
        else:
            ans[j] = x
            j += 2
    return ans
assert f([1, -2, 3, -4, 5, -6]) == ??
[/PYTHON]
[RAW_COT]
[THOUGHT]
# THINKING
[/THOUGHT]
[ANSWER]
assert f([1, -2, 3, -4, 5, -6]) == [1, -2, 3, -4, 5, -6]
[/ANSWER]
[/RAW_COT]
\end{Verbatim}

\vspace{0.2em}
\noindent\rule{\linewidth}{0.4pt}
\vspace{0.2em}

\textbf{Output:}
\begin{Verbatim}[fontsize=\scriptsize,
                 breaklines=true,
                 breakanywhere=true,
                 breaksymbolleft={},
                 breaksymbolright={}]
[COT]
[THOUGHT]
Let's execute the code step by step:
1. The function f is called with nums = [1, -2, 3, -4, 5, -6].
2. ans = [0] * len(nums) = [0] * 6 = [0, 0, 0, 0, 0, 0]. i = 0, j = 1.
3. Loop over each element x in nums:
   - x = 1: 1 > 0 is True, so ans[0] = 1, i = 2. ans = [1, 0, 0, 0, 0, 0].
   - x = -2: -2 > 0 is False, so ans[1] = -2, j = 3. ans = [1, -2, 0, 0, 0, 0].
   - x = 3: 3 > 0 is True, so ans[2] = 3, i = 4. ans = [1, -2, 3, 0, 0, 0].
   - x = -4: -4 > 0 is False, so ans[3] = -4, j = 5. ans = [1, -2, 3, -4, 0, 0].
   - x = 5: 5 > 0 is True, so ans[4] = 5, i = 6. ans = [1, -2, 3, -4, 5, 0].
   - x = -6: -6 > 0 is False, so ans[5] = -6, j = 7. ans = [1, -2, 3, -4, 5, -6].
4. The loop ends. The function returns ans = [1, -2, 3, -4, 5, -6].
[/THOUGHT]
[ANSWER]
assert f([1, -2, 3, -4, 5, -6]) == [1, -2, 3, -4, 5, -6]
[/ANSWER]
[/COT]
\end{Verbatim}
\end{promptbox}
\caption{An example of extracting vanilla tracing.}
\label{figure:extract_cot}
\end{figure}

\begin{figure}[p]
\begin{promptbox}[An example of the code reasoning of CodeReasoner.]
\textbf{Input:}
\begin{Verbatim}[fontsize=\small,
                 breaklines=true,
                 breakanywhere=true,
                 breaksymbolleft={},
                 breaksymbolright={}]
[SYSTEM_PROMPT]
...
[/SYSTEM_PROMPT]
[INSTRUCTION]
...
[/INSTRUCTION]
[PYTHON]
def f(words, x):
    res = []
    for i, w in enumerate(words):
        if x in w:
            res.append(i)
    return res

assert f(['leet', 'code'], 'e') == ??
[/PYTHON]
[THOUGHT]
\end{Verbatim}

\vspace{0.2em}
\noindent\rule{\linewidth}{0.4pt}
\vspace{0.2em}

\textbf{Output:}
\begin{Verbatim}[fontsize=\small,
                 breaklines=true,
                 breakanywhere=true,
                 breaksymbolleft={},
                 breaksymbolright={}]
Let's execute the code step by step:

1. **Function Call**: `f(['leet', 'code'], 'e')`
2. **Initialization**:
   - `words = ['leet', 'code']`
   - `x = 'e'`
   - `res = []`

3. **Loop Iteration**:
   - **First Iteration** (`i=0`, `w='leet'`):
     - Check if `x` ('e') is in `w` ('leet'): `'e' in 'leet'` evaluates to `True`.
     - Append `i` (0) to `res`: `res = [0]`.

   - **Second Iteration** (`i=1`, `w='code'`):
     - Check if `x` ('e') is in `w` ('code'): `'e' in 'code'` evaluates to `True`.
     - Append `i` (1) to `res`: `res = [0, 1]`.

4. **Return Statement**: After processing all elements, return `res = [0, 1]`.

[ANSWER]
assert f(['leet', 'code'], 'e') == [0, 1]
[/ANSWER]
\end{Verbatim}
\end{promptbox}
\caption{An example of the code reasoning of CodeReasoner. \textbf{\texttt{[correct]}}}
\label{figure:codereasoner_example}
\end{figure}

\begin{figure}[p]
\begin{promptbox}[An example of the code reasoning of ExecVerify.]
\textbf{Input:}
\begin{Verbatim}[fontsize=\small,
                 breaklines=true,
                 breakanywhere=true,
                 breaksymbolleft={},
                 breaksymbolright={}]
[SYSTEM_PROMPT]
...
[/SYSTEM_PROMPT]
[INSTRUCTION]
...
[/INSTRUCTION]
[PYTHON]
def f(words, x):
    l = []
    for i in range(len(words)):
        if x in words[i]:
            l.append(i)
    return l

assert f(['leet', 'code'], 'e') == ??
[/PYTHON]
[THOUGHT]
\end{Verbatim}

\vspace{0.2em}
\noindent\rule{\linewidth}{0.4pt}
\vspace{0.2em}

\textbf{Output:}
\begin{Verbatim}[fontsize=\small,
                 breaklines=true,
                 breakanywhere=true,
                 breaksymbolleft={},
                 breaksymbolright={}]
Let's execute the code step by step:

1. Function `f` is called with `words = ['leet', 'code']` and `x = 'e'`.
2. Initialize `l = []`.
3. Iterate over `range(len(words))`, which is `range(2)` (indices 0 and 1):

   - **Iteration 1 (i = 0)**:
     - Check if `'e' in words[0]`: `words[0]` is `'leet'`. Since `'e'` is in `'leet'`, append `0` to `l`. Now `l = [0]`.

   - **Iteration 2 (i = 1)**:
     - Check if `'e' in words[1]`: `words[1]` is `'code'`. Since `'e'` is in `'code'`, append `1` to `l`. Now `l = [0, 1]`.

4. Return `l`, which is `[0, 1]`.

[ANSWER]
assert f(['leet', 'code'], 'e') == [0, 1]
[/ANSWER]
\end{Verbatim}
\end{promptbox}
\caption{An example of the code reasoning of ExecVerify. \textbf{\texttt{[correct]}}}
\label{figure:execverify_example}
\end{figure}

\begin{figure}[p]
\begin{promptbox}[An example of the code reasoning of CWM.]
\textbf{Input:}
\begin{Verbatim}[fontsize=\small,
                 breaklines=true,
                 breakanywhere=true,
                 breaksymbolleft={},
                 breaksymbolright={}]
<|begin_of_text|><|trace_context_start|>
def f(words, x):
    res = []
    for i, w in enumerate(words):
        if x in w:
            res.append(i)
    return res


def main(): # << START_OF_TRACE
    return f(['leet', 'code'], 'e')
<|frame_sep|><|call_sep|>{}<|action_sep|>def main():
<|frame_sep|>
\end{Verbatim}

\vspace{0.2em}
\noindent\rule{\linewidth}{0.4pt}
\vspace{0.2em}

\textbf{Output:}
\begin{Verbatim}[fontsize=\small,
                 breaklines=true,
                 breakanywhere=true,
                 breaksymbolleft={},
                 breaksymbolright={}]
<|line_sep|>{}<|action_sep|>    return f(['leet', 'code'], 'e')
<|frame_sep|><|call_sep|>{"words": "['leet', 'code']", "x": "'e'"}<|action_sep|>def f(words, x):
<|frame_sep|><|line_sep|>{"words": "..", "x": ".."}<|action_sep|>    res = []
<|frame_sep|><|line_sep|>{"words": "..", "x": "..", "res": "[]"}<|action_sep|>    for i, w in enumerate(words):
<|frame_sep|><|line_sep|>{"words": "..", "x": "..", "res": "..", "i": "0", "w": "'leet'"}<|action_sep|>        if x in w:
<|frame_sep|><|line_sep|>{"words": "..", "x": "..", "res": "..", "i": "..", "w": ".."}<|action_sep|>            res.append(i)
<|frame_sep|><|line_sep|>{"words": "..", "x": "..", "res": "[0]", "i": "..", "w": ".."}<|action_sep|>    for i, w in enumerate(words):
<|frame_sep|><|line_sep|>{"words": "..", "x": "..", "res": "..", "i": "1", "w": "'code'"}<|action_sep|>        if x in w:
<|frame_sep|><|line_sep|>{"words": "..", "x": "..", "res": "..", "i": "..", "w": ".."}<|action_sep|>            res.append(i)
<|frame_sep|><|line_sep|>{"words": "..", "x": "..", "res": "[0, 1]", "i": "..", "w": ".."}<|action_sep|>    for i, w in enumerate(words):
<|frame_sep|><|line_sep|>{"words": "..", "x": "..", "res": "..", "i": "..", "w": ".."}<|action_sep|>    return res
<|frame_sep|><|return_sep|><|action_sep|>    return res
<|arg_sep|>"[0, 1]"<|frame_sep|><|return_sep|><|action_sep|>    return f(['leet', 'code'], 'e')
<|arg_sep|>"[0, 1]"<|frame_sep|><|end_of_text|>
\end{Verbatim}
\end{promptbox}
\caption{An example of the code reasoning of CWM. \textbf{\texttt{[correct]}}}
\label{figure:cwm_example}
\end{figure}


\end{document}